\documentclass{article}

\PassOptionsToPackage{numbers, compress}{natbib}
\usepackage[preprint]{neurips_2022}

\usepackage[utf8]{inputenc} % allow utf-8 input
\usepackage[T1]{fontenc}    % use 8-bit T1 fonts
\usepackage{hyperref}       % hyperlinks
\usepackage{url}            % simple URL typesetting
\usepackage{booktabs}       % professional-quality tables
\usepackage{amsfonts}       % blackboard math symbols
\usepackage{nicefrac}       % compact symbols for 1/2, etc.
\usepackage{microtype}      % microtypography
\usepackage{amsmath}
\usepackage{textcomp}
\usepackage{arydshln}
\usepackage{caption}
\usepackage{subcaption}
\usepackage{graphicx}
\usepackage{algorithm}
\usepackage{algorithmic}
\usepackage{wrapfig}

\title{Time-series Transformer Generative Adversarial Networks}

\author{%
  \textbf{Padmanaba Srinivasan} \\
  Department of Computing\\
  Imperial College London\\
  London, SW7 2AZ, UK \\
  \texttt{ps3416@ic.ac.uk} \\
  \And
  \textbf{William J. Knottenbelt} \\
  Department of Computing\\
  Imperial College London\\
  London, SW7 2AZ, UK \\
  \texttt{wjk@ic.ac.uk} \\
}

\begin{document}

\maketitle

\begin{abstract}
    Many real-world tasks are plagued by limitations on data: in some instances very little data is available and in others, data is protected by privacy enforcing regulations (e.g.\ GDPR). We consider limitations posed specifically on time-series data and present a model that can generate synthetic time-series which can be used in place of real data. A model that generates synthetic time-series data has two objectives: 1) to capture the stepwise conditional distribution of real sequences, and 2) to faithfully model the joint distribution of entire real sequences. Autoregressive models trained via maximum likelihood estimation can be used in a system where previous predictions are fed back in and used to predict future ones; in such models, errors can accrue over time. Furthermore, a plausible initial value is required making MLE based models not really \textit{generative}. Many downstream tasks learn to model conditional distributions of the time-series, hence, synthetic data drawn from a generative model must satisfy 1) in addition to performing 2). We present TsT-GAN, a framework that capitalises on the Transformer architecture to satisfy the desiderata and compare its performance against five state-of-the-art models on five datasets and show that TsT-GAN achieves higher predictive performance on \textbf{all} datasets.
\end{abstract}

\section{Introduction}

Many real world applications are reliant on time-series data, however, not all these applications necessarily have available the data they need. For some tasks, data is available only in small quantities -- often too small to base detailed analysis on -- and for others, data is protected by regulatory or ethical concerns . Medicine is one field that is notorious for suffering from such problems and the ability to generate synthetic data is an avenue through which analysis can continue \cite{RN617,RN618,RN619,RN620}. Another field with similar issues is data from human-internet interactions; as world governments have increasingly adopted GDPR-like legislation, the availability of such data is limited and methods that can generate synthetic data as a stand-in for user data will increase in importance. Although we do not expressly use privacy-preserving methods in training, this property can be achieved using methods such as differential privacy-preserving stochastic gradient descent \cite{RN621,RN622}. 

Good quality synthetic time-series data should respect the conditional distribution of a time-series $\prod_{t=1}^{T}p(\mathbf{x}_t \mid \mathbf{x}_1, ..., \mathbf{x}_t)$, so as to maximise utility for downstream models. Generated synthetic data should also capture well the joint distribution $p(\mathbf{x}_1, ..., \mathbf{x}_T)$ such that the synthetic data is indistinguishable from real data. 

A straightforward, if na\"ive approach to generating synthetic data is to use autoregressive models trained using teacher forcing \cite{RN462,RN601} by repeatedly feeding it past predictions. Conditioning on previous outputs is prone to errors adding up over the course of a sequence and techniques to overcome this \cite{RN584,RN602,RN585} have not entirely solved the problem \cite{RN510}.

Generative models are statistical models that learn from a set of data instances, \textit{X}, and their corresponding labels, \textit{Y} to capture the joint probability distribution $p(X, Y)$. Learning the joint distribution allows generative models to generate new data instances. Generative Adversarial Networks (GANs) are one method of training generative models \cite{RN458} that typically map some noise, \textit{z}, to a posterior distribution. Specifically, for time-series, GANs incorporate an additional temporal dimension to model the joint distribution of all elements of the time-series. Synthetically generated sequences have a wide array of applications \cite{RN603,RN604,RN605}, yet in many existing approaches the true time-series dynamics are not explicitly learned and the quality of how well these dynamics are learned are not explored in detail.

%Summary of results,
%test across n time-series + sttement of results

\textbf{Contributions} We use the two characteristics of good generative models to guide the development of a new architecture that that contains a generator that can model the full joint distribution while also respecting the need for accurate conditional distributions. We develop a training framework for our model that can be applied to any time-series dataset and benchmark our method quantitatively using the \textit{train on synthetic test on real} (TS-TR) \cite{RN559} approach, and qualitatively using t-SNE \cite{RN591}. We compare our model against five state-of-the-art baselines on five datasets and show that TsT-GAN consistently achieves superior performance, achieving the best predictive scores across the board while also demonstrating best discriminative performance on three out of five datasets. We also perform ablation experiments on TsT-GAN to identify sources of performance gain.

\section{Related Work} 

\textbf{Generative Adversarial Networks} A GAN model consists of two components: firstly, a generator that maps noise to a posterior distribution; and a discriminator whose job it is to distinguish between samples produced by the generator and samples drawn from the dataset. Some popular applications of GANs focus on generating high-dimensional data, such as images \cite{RN552,RN551,RN550} and videos \cite{RN555,RN554,RN556}. Developments have also been made to make the notoriously unstable process of training GANs more stable \cite{RN552,RN509,RN508,RN553,RN518}.

The output of a generator can be controlled by conditioning on additional information. This class of GANs are called conditional GANs (cGANS) \cite{RN497} where the generator takes as an additional input some information to direct the generative process. cGANs have been applied to generate sequential data in a number of fields, such as, video clips \cite{RN554,RN555,RN556}, time-series data \cite{RN559,RN560,RN510}, natural language tasks \cite{RN576,RN575,RN577} and generating tabular data \cite{RN582,RN580,RN581,RN579}. Generating new data is a task well suited to GANs -- the generator is trained to generate data by learning the underlying generating distribution and guided by the discriminator. Once trained, a generator can be used to generate synthetic data samples for cases where limited real data is available.

\textbf{Time-Series Generative Models} Generating synthetic time-series data extends the generation of synthetic tabular data by incorporating an additional temporal dependency. This tasks generative models with learning features of the data within each time step and also relating features across time.

One approach to designing generative models for time-series generation is Professor Forcing \cite{RN585} which combines a GAN framework with a supervised learning approach where a Recurrent Neural Network (RNN) based generator \cite{RN384,RN610} alternates between teacher forcing \cite{RN462,RN601} and generative training and the discriminator distinguishes between hidden states produced in teacher forcing mode and free running mode. This encourages the generator to match the conditional distributions between the two modes. 

The C-RNN-GAN framework \cite{RN560} directly applies Long short-term memory (LSTM) \cite{RN462} neural network in both generator and discriminator to generate sequential data. The generator receives noise inputs at each time step and generates some data, conditioned on previous outputs. RCGAN \cite{RN559} extends this approach by allowing conditioning on additional information while also removing the dependence on previous outputs.

TimeGAN \cite{RN510} modifies the standard GAN framework and adopts aspects of Professor Forcing. The framework uses four components: an embedder and decoder (trained via teacher forcing as a joint autoencoder network), a generator and a discriminator. The generator receives noise input and produces hidden states which are passed to the discriminator along with the hidden states of the embedder, which discriminates between the embedder and generator latent distributions. An additional supervised loss penalises differences between the two distributions. COT-GAN \cite{RN592} presents a flexible GAN architecture trained using a novel adversarial loss that builds on the Sinkhorn divergence \cite{RN593}.

\section{Method}

\subsection{Problem Formulation}
\label{subsection: problem formulation}

We denote a multivariate sequence of length \textit{T} as $\mathbf{x}_{1:T} = \mathbf{x}_1, ..., \mathbf{x}_T$ with \textit{N} such sequences form the training dataset $\mathcal{D} = \{\mathbf{x}_{1:T}\}_{n=1}^N$. The aim of the generator is to learn a distribution $\hat{p}(\mathbf{x}_{1:T})$ as an approximation to the generating distribution $p(\mathbf{x}_{1:T})$. Learning the conditional distribution $\prod_{t} p(\mathbf{x}_t \mid \mathbf{x}_1, ..., \mathbf{x}_{t-1})$ is a far simpler objective learned by single time-step autoregressive models. 

Any synthetic data generated using a trained generator is likely to be used downstream to train autoregressive models. As a result, in addition to learning the joint distribution $p(\mathbf{x}_{1:T})$ the generator must also learn $\prod_{t} p(\mathbf{x}_t \mid \mathbf{x}_1, ..., \mathbf{x}_{t-1})$. The joint distribution suggests that the generative model can be bidirectional, whereas an autoregressive constraint is explicit in the conditional distribution. To this end, we propose two objectives. The first is to learn the joint distribution:

\begin{equation}
    \label{eq:global objective}
    \underset{\hat{p}}{\mathrm{min}}\ \ D( p(\mathbf{x}_{1:T}) \mid\mid \hat{p}(\mathbf{x}_{1:T}) )
\end{equation}

Which we interpret as a global objective where the learned joint distribution must match the true joint distribution. We also incorporate the conditional distribution:

\begin{equation}
    \label{eq:stepwise objective}
    \underset{\hat{p}}{\mathrm{min}}\ \ D( p(\mathbf{x}_t \mid \mathbf{x}_{1:t-1}) \mid\mid \hat{p}(\mathbf{x}_t \mid \mathbf{x}_{1:t-1}) )
\end{equation}

where \textit{D} represents suitable distance metrics for the two objectives. This represents a local (autoregressive) objective where each item is conditioned on previous ones.

\subsection{Proposed Model}

We present our TsT-GAN model which consists of four components, each designed with respect to the objectives in Section~\ref{subsection: problem formulation}.

\subsubsection{Embedder--Predictor}
\label{subsubsection: embedder-predictor}

The embedder--predictor network consists of a transformer network that takes as input real multivariate sequences $\mathbf{x}_{1:T}$ and predicts the next item in the sequence at each position. This network consists of a linear projection of the input vector into the model dimension. The projected sequence is passed through the embedder network $\textit{E}_\theta$ to produce the set of final embeddings $\hat{\mathbf{h}}_1, ..., \hat{\mathbf{h}}_T$. The predictor network $\textit{P}_\theta$ takes the embeddings back into the original input dimension $\textit{P}_\theta: \mathbb{R}^d \rightarrow \mathbb{R}^m$ and is implemented by a separate neural network. Applying the predictor network to all the embeddings produced by the embedding network generated one step ahead predictions for all positions in the input sequence $\hat{\mathbf{x}}_1, ..., \hat{\mathbf{x}}_T$. 

The embedding network is parameterised by a transformer decoder network, which uses an autoregressive mask and we realise the predictor network as a simple linear layer that performs a linear projection of the embedding back into the original data input dimension.

The embedder--predictor network is trained using a supervised loss that penalises one step ahead prediction error:

\begin{equation}
    \mathcal{L}_S (\mathbf{x}_{1:T}, \hat{\mathbf{x}}_{1:T}) = 
    \frac{1}{T-1} \sum_{t=1}^{T-1} \lvert\lvert \mathbf{x}_{t+1} - \hat{\mathbf{x}}_t \rvert\rvert_2
\end{equation}

where $\hat{\mathbf{x}}_t$ denotes the prediction by the embedder--predictor network at position \textit{t} and that corresponds to the predicted value of the time-series at time \textit{t+1}. This allows learning of the true conditional distribution $\prod_t p(\mathbf{x}_t \mid \mathbf{x}_{1:t-1})$ with the embedder modelling the latent conditional distribution $\prod_t p(\hat{\mathbf{h}}_t \mid \hat{\mathbf{h}}_{1:t-1})$.

\subsubsection{Generator and Discriminator}

The generator model $\textit{G}_\theta$ takes a sequence of random vectors $\mathbf{z}_1, ..., \mathbf{z}_T$ and projects these into the model dimension. The projected noise vector is passed through $\textit{G}_\theta$ which outputs a set of latent embeddings $\tilde{\mathbf{h}}_1, ..., \tilde{\mathbf{h}}_T$. Each latent embedding $\tilde{\mathbf{h}}_t \in \mathbb{R}^d$ is then transformed back into the original input space by way of the predictor network from Section~\ref{subsubsection: embedder-predictor} to produce a synthetic sequence $\tilde{\mathbf{x}}_{1:T} = \tilde{\mathbf{x}}_1, ..., \tilde{\mathbf{x}}_T$. Parameters of the predictor network are shared between the generator and embedder. 

We construct $\textit{G}_\theta$ in a similar way to the embedding network; it consists of a transformer encoder that makes use of bidirectional attention. To enforce the autoregressive property, we allow the parameters of the predictor network to be updated only when performing backpropagation through the embedder--predictor network. When backpropagating through the generator-predictor network, gradients are calculated but the parameters of the predictor network are frozen. This forces the generator to learn the latent conditional distributions of the embedder to produce valid synthetic data while also allowing full treatment of the joint probability. The random vectors fed into the generator can be drawn from any distribution, we draw random vectors from a standard Gaussian distribution.

The discriminator model $\textit{D}_\theta$ is constructed in a similar way to BERT \cite{RN457} as a transformer encoder with bidirectional attention. A linear projection is used to map input sequences to the model dimension following which a \texttt{[CLS]} embedding is prepended to the beginning of the sequence. This sequence is passed through the discriminator and the embedding corresponding to the \texttt{[CLS]} position is projected into $\mathbb{R}^1$ for classification. The discriminator receives as input real sequences drawn from the dataset, which it is tasked with classifying as \textit{true}, and synthetic sequences from the generator, which it must classify as \textit{false}. Our discriminator design focuses on a global classification of the quality of a sequence, which differs from previous RNN based approaches which classify on a per time step basis. By performing global sequence classification with the discriminator, we address our first objective in Equation~\ref{eq:global objective}, while the stepwise objective in Equation~\ref{eq:stepwise objective} is handled indirectly via the embedder--predictor system. We apply the LS-GAN adversarial loss \cite{RN508}, which uses separate objectives for the discriminator and generator:

\iffalse
\begin{equation}
    \label{eq:gan loss}
    \mathcal{L}_{GAN} =
    \underset{\textit{G}_\theta}{\mathrm{min}}\ \underset{\textit{D}_\theta}{\mathrm{max}}\ \
    \frac{1}{2} \mathbb{E}_{\mathbf{x}_{1:T} \sim p} [\log(D_\theta(\mathbf{x}_{1:T})] + \frac{1}{2}\mathbb{E}_{\tilde{\mathbf{x}}_{1:T} \sim \hat{p}} [\log(1 - D_\theta(\tilde{\mathbf{x}}_{1:T})]
\end{equation}
\fi

\begin{align}
\begin{split}
    \label{eq:gan loss}
    \mathcal{L}_{GAN}(D_\theta) =
    \underset{\textit{D}_\theta}{\mathrm{min}}\ \
    \mathbb{E}_{\mathbf{x}_{1:T} \sim p} [((D_\theta(\mathbf{x}_{1:T}) - 1)^2] + \mathbb{E}_{\tilde{\mathbf{x}}_{1:T} \sim \hat{p}} [(D_\theta(\tilde{\mathbf{x}}_{1:T})^2]
    \\
    \mathcal{L}_{GAN}(G_\theta) =
    \underset{\textit{G}_\theta}{\mathrm{min}}\ \
    \frac{1}{2} \mathbb{E}_{\tilde{\mathbf{x}}_{1:T} \sim \hat{p}} [(D_\theta(\tilde{\mathbf{x}}_{1:T})-1)^2]
\end{split}
\end{align}

The GAN objective is likely to be insufficient to fully capture the temporal dependencies across long time periods \cite{RN608}. Applying a second, supervised autoregressive loss on the generator is not possible as the generator is bidirectional, so we turn to unsupervised masked training. Following a similar method to masked language modelling (MLM) in BERT \cite{RN457}, we randomly mask out items in a sequence with probability $p_{mask}$. Masked out positions are replaced with a learnable \texttt{[MASK]} embedding and the generator is tasked with predicting the true values at the masked positions. Transformers have been applied to autoregressive time-series tasks \cite{RN455,RN456,RN589} and have been shown to benefit from masked modelling pre-training \cite{RN455}. We add the following masked modelling objective:

\begin{equation}
    \label{eq:masked loss}
    \mathcal{L}_{MM} (\mathbf{x}_{1:T}, \overline{\mathbf{x}}_{1:T}) = 
    \frac{1}{\lvert M \rvert} \sum_{t \in M} \lvert\lvert \mathbf{x}_t - \overline{\mathbf{x}}_t \rvert\rvert_2
\end{equation}

where \textit{M} denotes the masked positions and $\overline{\mathbf{x}}_t$ is the output of the generator at position \textit{t} when performing masked modelling. We perform full generation and masked modelling in an alternating fashion using separate learnable linear projections for both tasks.

\subsection{Architecture Overview}

\begin{figure}
    \centering
    \includegraphics[trim=4cm 3cm 4cm 3cm, clip, width=0.6\textwidth]{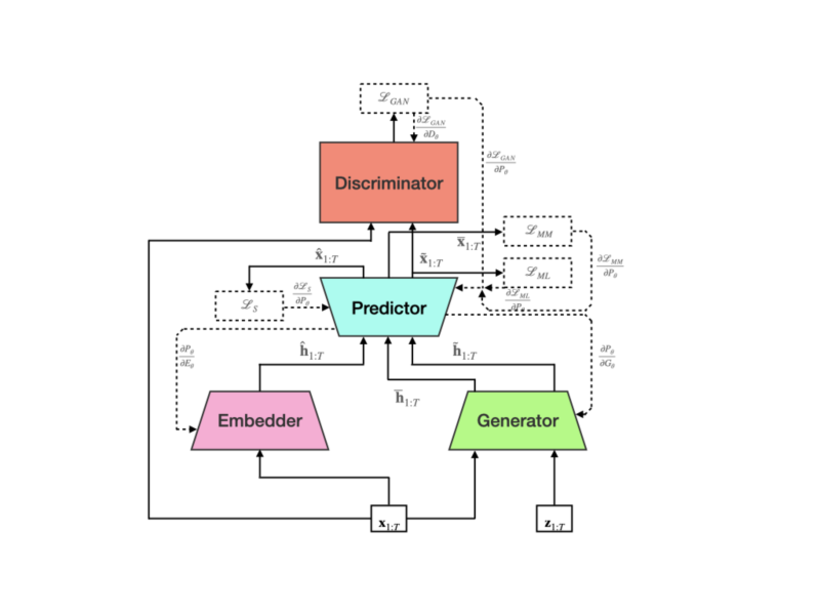}
    \caption{Block diagram of TsT-GAN showing data and gradient flows. Gradients that propagate back to the generator will pass through the predictor network, but, will not be allowed to change the parameters of the predictor. Predictor parameters are only updated with respect to gradient $\frac{\partial \mathcal{L}_S}{\partial \textit{P}_\theta}$.}
    \label{fig:block diagram}
\end{figure}

Given that $\mathbf{x}_{1:T} = \mathbf{x}_1, ..., \mathbf{x}_T$ denotes a true time-series of length \textit{T} and $\mathbf{z}_{1:T} = \mathbf{z}_1, ..., \mathbf{z}_T$ denotes a sequence of random vectors drawn from an isotropic Gaussian, we provide formal motivation for our model, with an accompanying block diagram in Figure~\ref{fig:block diagram}. The embedder, trained using maximum likelihood estimation (MLE) takes as input $\mathbf{x}_{1:T}$ and produces conditional latent embeddings $\hat{\mathbf{h}}_{1:T} = \hat{\mathbf{h}}_1, ..., \hat{\mathbf{h}}_T$ which the predictor maps to values $\hat{\mathbf{x}}_{1:T}$. 

The generator takes as input a sequence of random vectors $\mathbf{z}_{1:T}$ and produces a corresponding set of latent embeddings $\tilde{\mathbf{h}}_{1:T} =  \tilde{\mathbf{h}}_1, ..., \tilde{\mathbf{h}}_T$. To maximise the downstream utility of synthetic data, the generator aims to learn the conditional latent distribution produced by the embedder such that $p(\tilde{\mathbf{h}}_t) \approx p(\hat{\mathbf{h}}_t)\ \forall t \in T$. We achieve this by sharing predictor parameters between the generator and discriminator, but allowing the predictor's parameters to be changed only when updating with respect to the gradient $\frac{\partial \mathcal{L}_S}{\partial \textit{P}_\theta}$, thereby forcing the condition.

%\begin{wrapfigure}{r}{0.45\textwidth}
%\end{wrapfigure}

The discriminator is tasked with differentiating between real sequences $\mathbf{x}_{1:T}$ and synthetic sequences $\tilde{\mathbf{x}}_{1:T}$. The discriminator operates over entire sequences, producing only one true/false classification and so, inspects synthetic sequences on a global scale and encourages the generator to learn the joint distribution of entire sequences. The objective $\mathcal{L}_{MM}$ reinforces this by while also exposing the generator to real samples and further encouraging bidirectional learning of the joint distribution. 

Some approaches to time-series generation \cite{RN608} have shown that explicit moment matching can improve the quality of synthetic data. We introduce an auxiliary moment loss to promote matching of first and second moments:

\begin{equation}
    \mathcal{L}_{ML} (\mathbf{x}_{1:T}, \tilde{\mathbf{x}}_{1:T})= \lvert f_\mu(\mathbf{x}_{1:T}) - f_\mu(\tilde{\mathbf{x}}_{1:T}) \rvert + \lvert f_\sigma(\mathbf{x}_{1:T}) - f_\sigma(\tilde{\mathbf{x}}_{1:T}) \rvert
\end{equation}

where $f_\mu$ and $f_\sigma$ are functions that compute the mean and standard deviation of a time-series.

\subsection{Optimisation}
\label{subsec: optimisation}

We train our model in three stages. We begin by training the embedder--predictor components independently using the objective $\mathcal{L}_S$, followed by training the generator using only the masked modelling objective $\mathcal{L}_{MM}$. The final stage consists of joint training using all objectives.

All of our transformer model components are feed forward architectures that are insensitive to ordering of sequence. Hence, after the initial projection to the model dimension we add sinusoidal position embeddings to projected embeddings. For all transformer components, we use a model dimension of $d = 32$, with $H = 8$ attention heads and a hidden layer dimension of $D = 4 \times d = 128$ with 3 encoder layers for each component. We use the GELU activation function \cite{RN463} for the non-linearity with LayerNorm \cite{RN606} normalisation.

For optimisation, we use the Adam optimiser \cite{RN484}, with a learning rate of $0.001$ in the first two training stages for the embedder, predictor and generator, followed by a learning rate of $0.00002$ for all components during joint training, with betas of $(0.5, 0.999)$. For masked modelling, we use $p_{mask} = 0.3$ for all datasets and a mini-batch size of 128.

\section{Experiments}

\subsection{Evaluation Methodology}

We compare the performance of our model on four different datasets against several baselines, including TimeGAN \cite{RN510}, RCGAN \cite{RN559}, C-RNN-GAN \cite{RN560}, COT-GAN \cite{RN592} and Professor Forcing (P-Forcing) \cite{RN585} models. We also perform ablation experiments, removing components of our TsT-GAN to identify sources of performance gain. 

We consider the following evaluation metrics:

\begin{enumerate}
    \item \textbf{Predictive Score} We generate a synthetic dataset using trained a generative model and train a post-hoc network on the synthetic data, after which we evaluate the post-hoc regression network on real data. If a generative model has captured the conditional distribution correctly, then we expect test mean absolute error (MAE) to be low and similar to when post-hoc network is trained on real data. The predictive score follows the TS-TR evaluation methodology \cite{RN559}. An ideal generator will produce synthetic samples that, under the TS-TR framework, will produce a predictive score no worse than when the post-hoc model is trained on real data.
    
    \item \textbf{Discriminative Score} A post-hoc network is trained to distinguish between real and synthetic data. The training set has an equal number of real and synthetic samples. We report the classification error on the held out test set. In the case of the ideal generator, synthetic samples are indistinguishable from real samples, resulting in a discriminative score of 0. 
    
    \item \textbf{Visualisation} We use t-SNE \cite{RN591} to reduce the dimensionality of real and synthetic datasets, flattening across the temporal dimension. This allows comparison of how well the synthetic data distribution matches the original, indicating any areas of the original distribution not captured as well as out of distribution samples in the synthetic data. In addition, to evaluate how well the joint distribution is captured, we calculate the first difference for real and synthetic time-series and plot t-SNEs.
\end{enumerate}

The code for each of the aforementioned models is available in public repositories published by the corresponding authors. We parameterise these models with all RNN-based components consisting of three layers of size 32. We use a sequence length of 24 for all datasets. Code associated with each of the models is used to train and generate synthetic data and evaluated using code similar to \cite{RN510}. For both the predictive and discriminative evaluations, we use a two layer Gated Recurrent Unit (GRU) \cite{RN610} with hidden size equal to the input dimension.

\subsection{Datasets}

We evaluate TsT-GAN across a range of datasets with different properties. All datasets we use are available online or can be generated.

\begin{enumerate}
    \item \textbf{Sines} The quality of sine waves can be evaluated easily by inspection and this dataset consists of a number of sine waves with random shifts in phase and frequency. Phase and frequency shifts are random variables: phase shift in $\phi \sim \textit{Uniform}[-\pi, \pi]$ and frequency shift $\lambda \sim \textit{Uniform}[0, 1]$. We generate a multivariate Sines dataset consisting of 5 sine waves per sample. This synthetic dataset provides continuous valued, periodic functions with no correlations between features. \textbf{(5 features and 10\ 000 rows.)}
    
    \item \textbf{Stocks} The Stocks dataset consists of daily data collected between 2004 and 2019 for the Google ticker.\footnote{Obtained from: \url{https://github.com/jsyoon0823/TimeGAN}} \textbf{(5 features and 3685 rows.)}
    
    \item \textbf{Energy} The UCI Appliances Energy Prediction dataset \cite{RN596, UCIRepo} is high-dimensional and consists of features such as energy consumption, humidity and temperature collected by sensors. The data is complex with samples logged every 10 minutes for around four and a half months. \textbf{(28 features and 19\ 735 rows.)}
    
    \item \textbf{Chickenpox} The UCI Hungarian Chickenpox Cases dataset \cite{RN595, UCIRepo} consists of records of chickenpox cases weekly in 20 counties in Hungary. This dataset represents a realistic situation where generative models be trained on small amounts of data and then generate synthetic samples to train other models. \textbf{(20 features and 521 rows.)}
    
    \item \textbf{Air} The UCI Air Quality dataset \cite{RN611,UCIRepo} consists of levels of different gases recorded hourly in an Italian city. We remove the date and time columns as part of preprocessing. \textbf{(13 features and 9358 rows.)}
\end{enumerate}

\subsection{Visualisation, Predictive and Discriminative Scores}

Table~\ref{tab:predictive discriminative scores} shows that TsT-GAN consistently creates more \textit{useful} data than the baseline models, achieving a lower predictive score across all datasets. Our predictive score for all datasets is remarkably close to the original score and the scores on the synthetic Sines and Stocks datasets outperform the original. TsT-GAN outperforms the next best performing baseline on the Sines, Energy and Stock datasets by 32\%, 22\% and 19\%, respectively. TsT-GAN performs consistently in the discriminative tests, achieving best performance across two datasets. COT-GAN achieves an incredible 0.6\% discriminative score on Chickenpox while exhibiting competitive predictive score performance with TsT-GAN. 

\begin{figure*}[t!]
    \centering
    
    \begin{subfigure}[t]{0.14\textwidth}
        \centering
        \includegraphics[width=\textwidth]{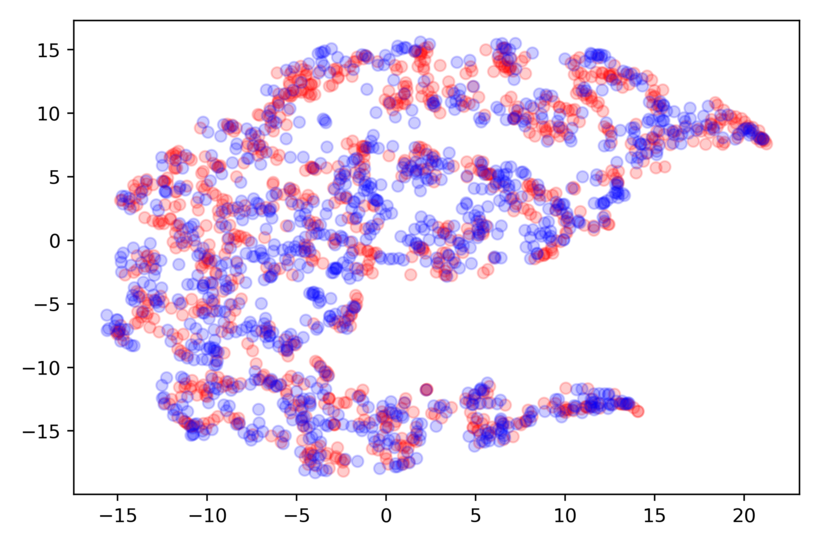}
        \includegraphics[width=\textwidth]{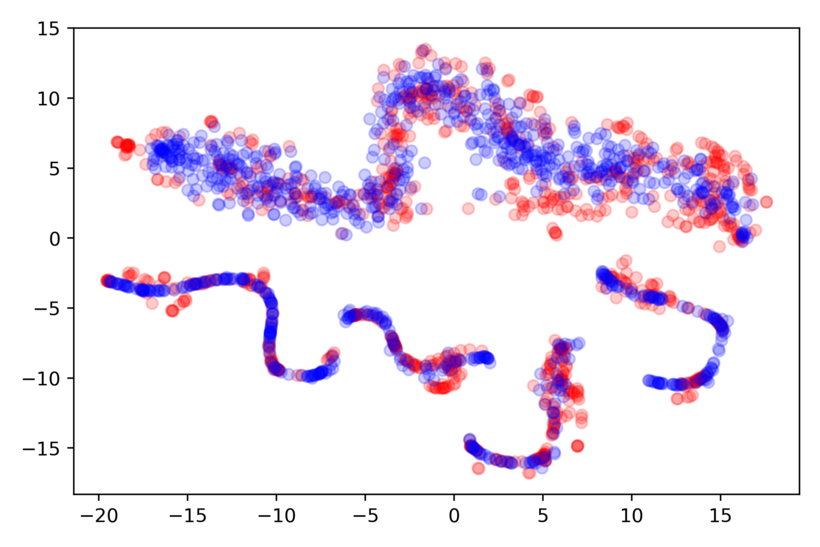}
        \includegraphics[width=\textwidth]{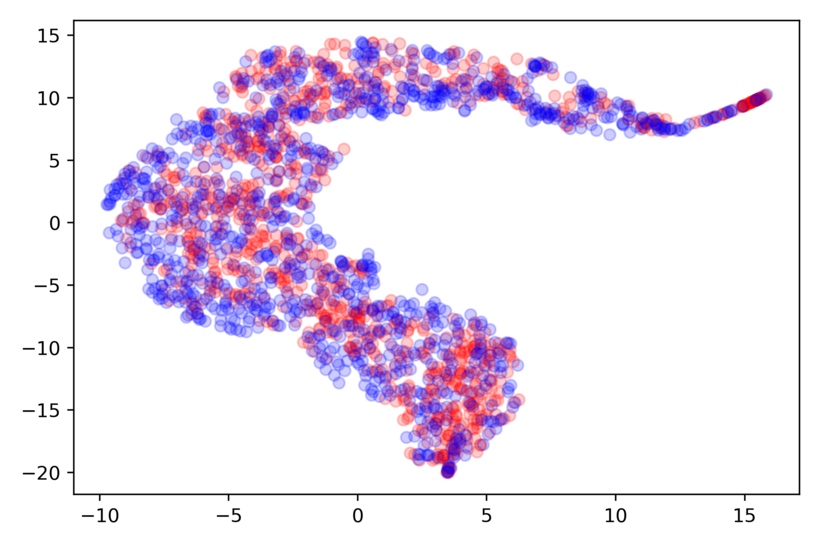}
        \includegraphics[width=\textwidth]{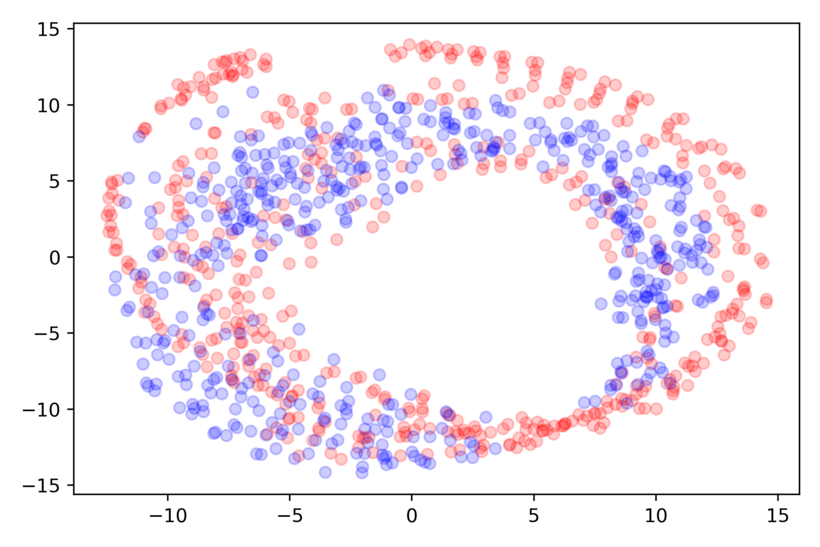}
        \includegraphics[width=\textwidth]{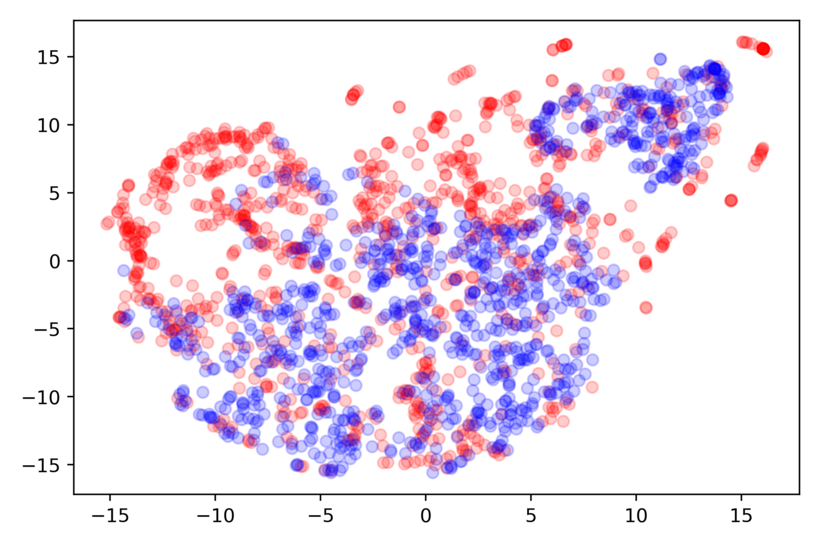}
        \caption{TsT-GAN}
    \end{subfigure}
    ~ 
    \begin{subfigure}[t]{0.14\textwidth}
        \centering
        \includegraphics[width=\textwidth]{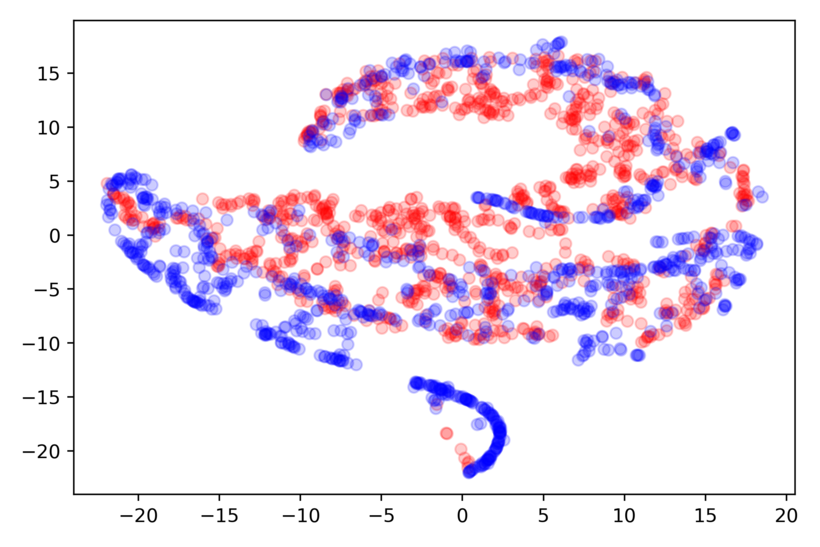}
        \includegraphics[width=\textwidth]{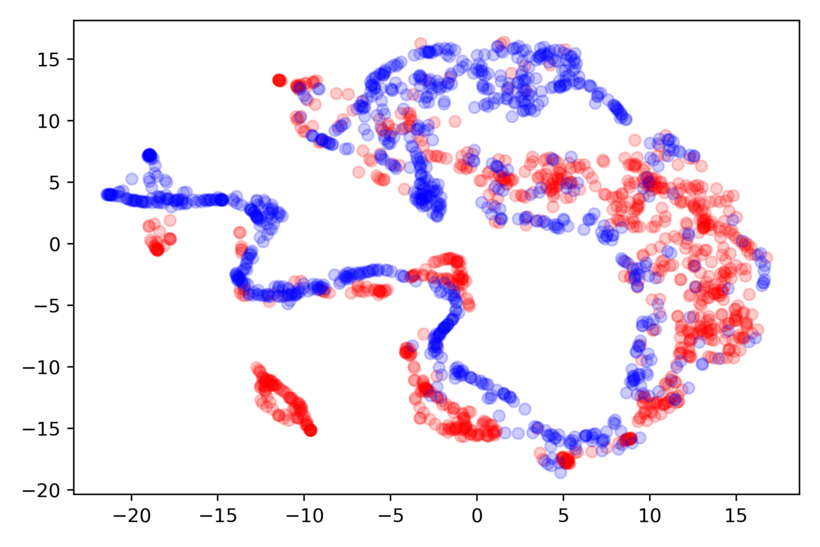}
        \includegraphics[width=\textwidth]{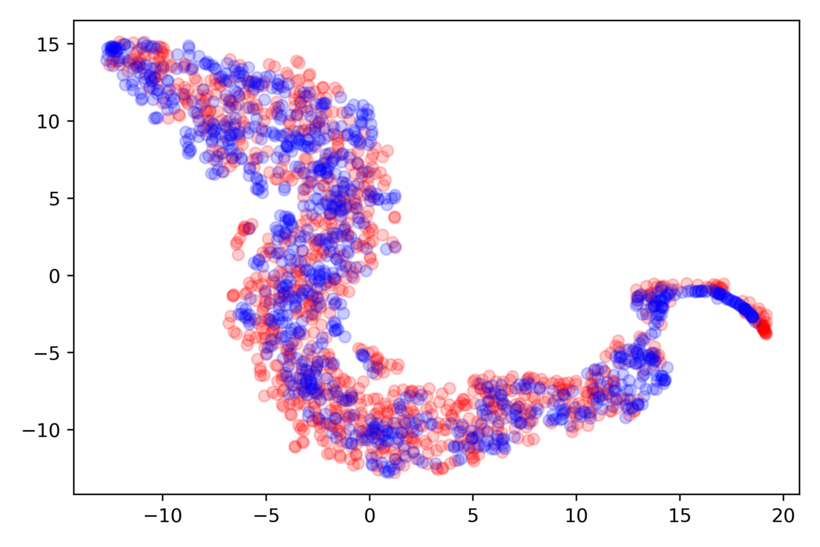}
        \includegraphics[width=\textwidth]{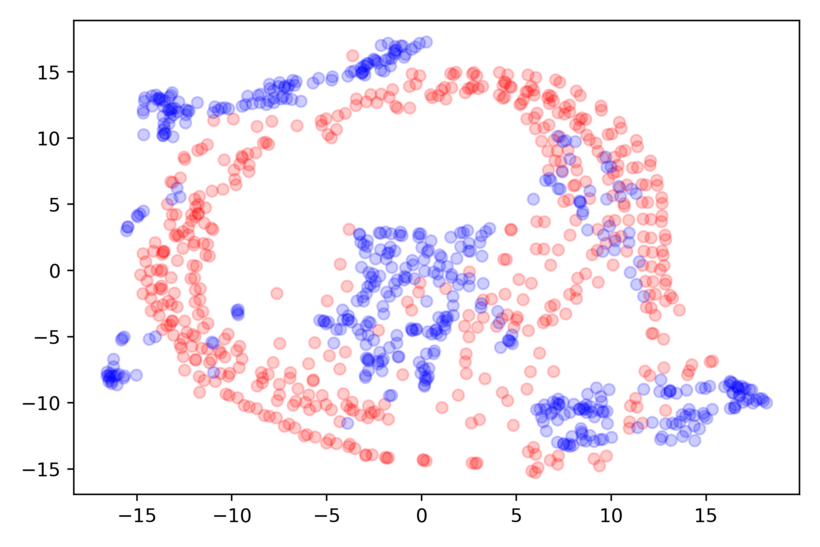}
        \includegraphics[width=\textwidth]{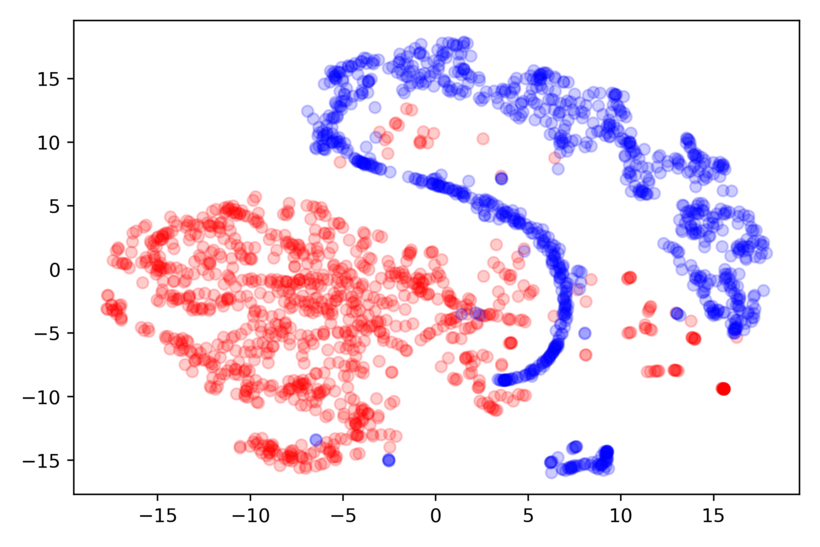}
        \caption{TimeGAN}
    \end{subfigure}
    ~
    \begin{subfigure}[t]{0.14\textwidth}
        \centering
        \includegraphics[width=\textwidth]{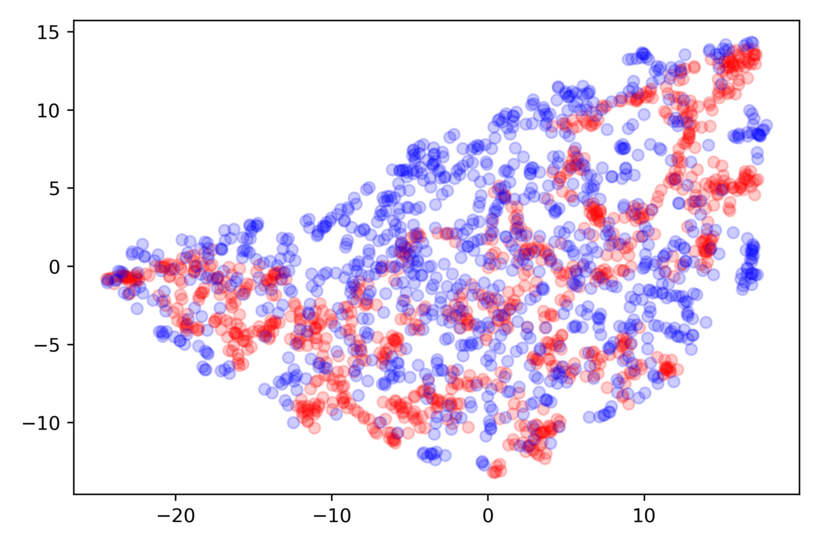}
        \includegraphics[width=\textwidth]{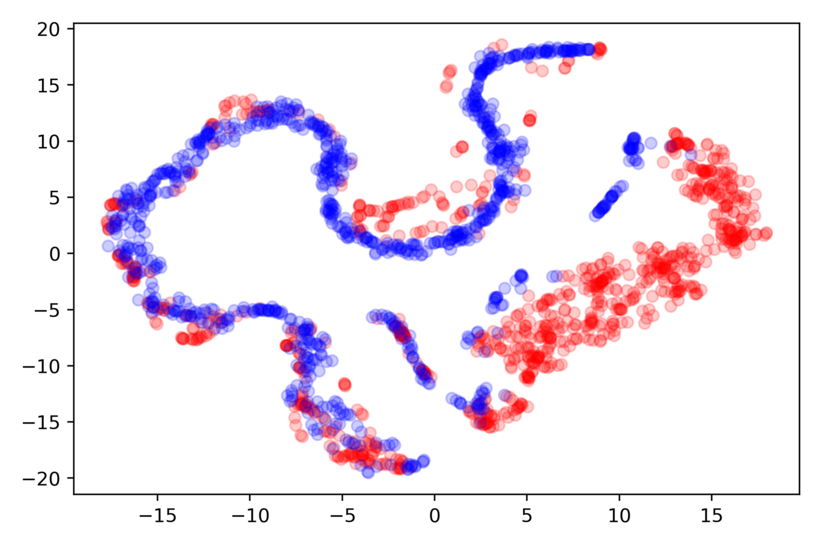}
        \includegraphics[width=\textwidth]{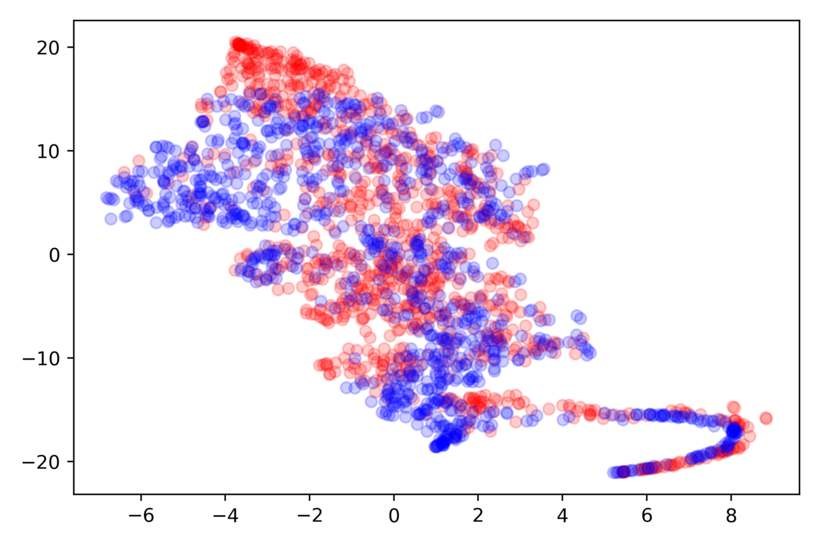}
        \includegraphics[width=\textwidth]{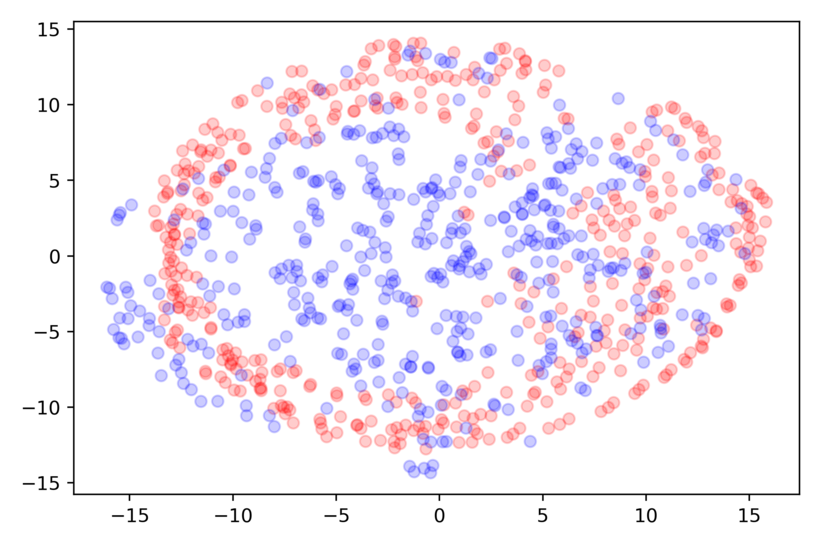}
        \includegraphics[width=\textwidth]{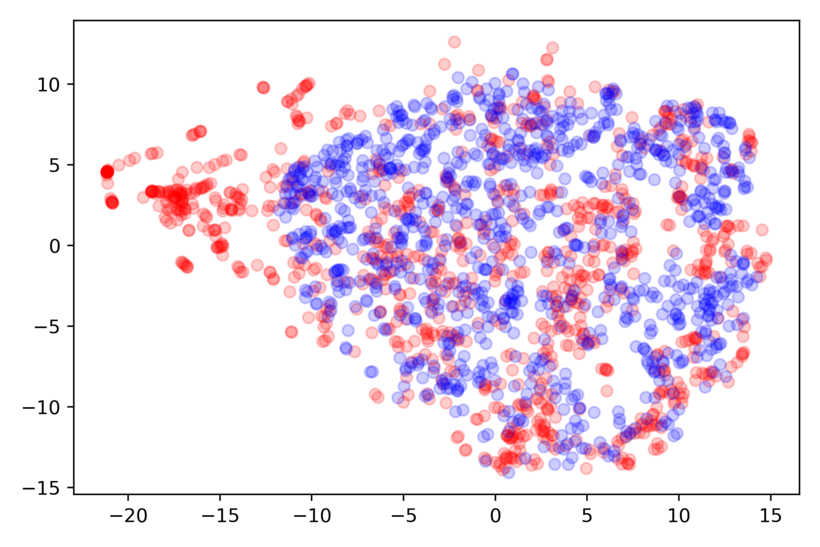}
        \caption{RCGAN}
    \end{subfigure}
    ~ 
    \begin{subfigure}[t]{0.14\textwidth}
        \centering
        \includegraphics[width=\textwidth]{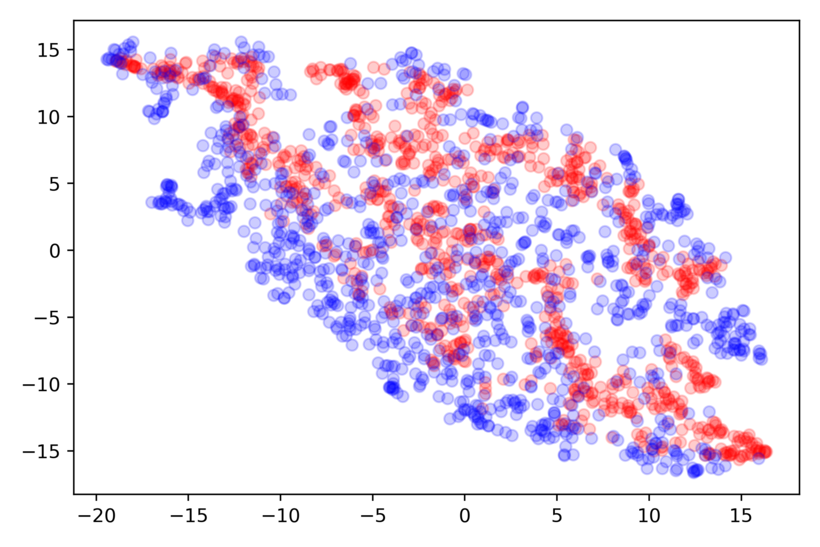}
        \includegraphics[width=\textwidth]{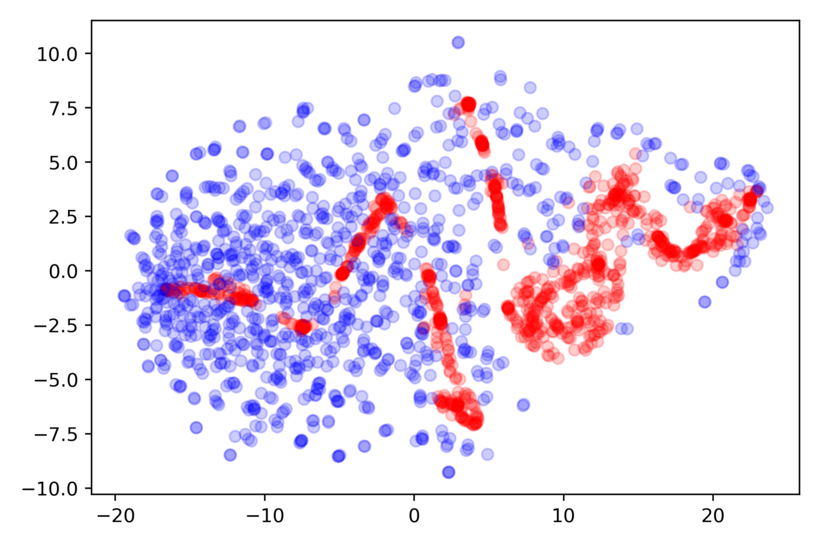}
        \includegraphics[width=\textwidth]{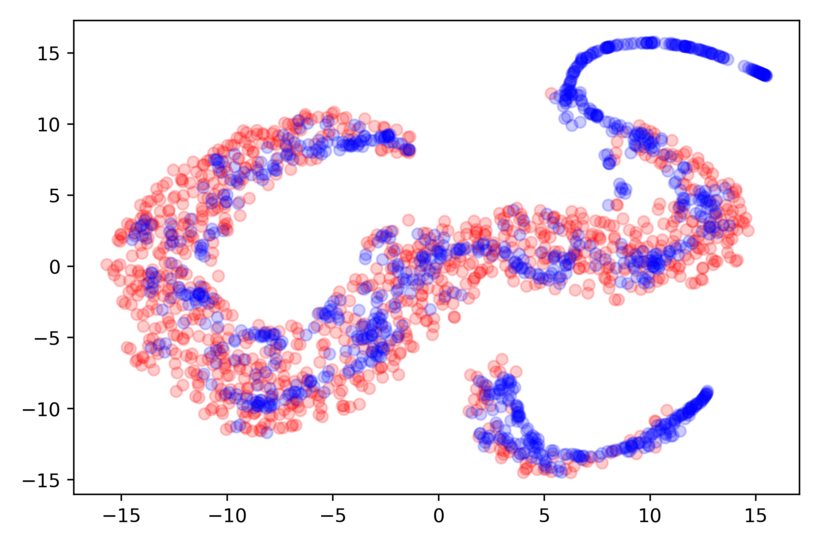}
        \includegraphics[width=\textwidth]{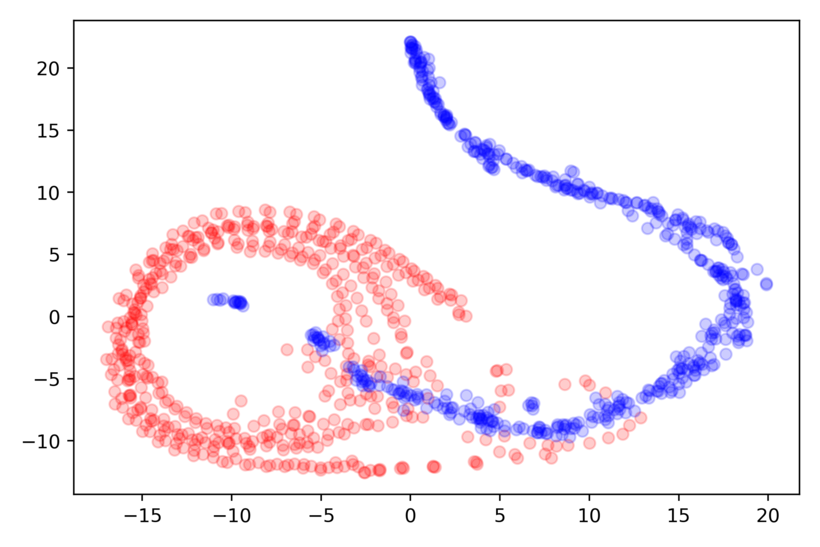}
        \includegraphics[width=\textwidth]{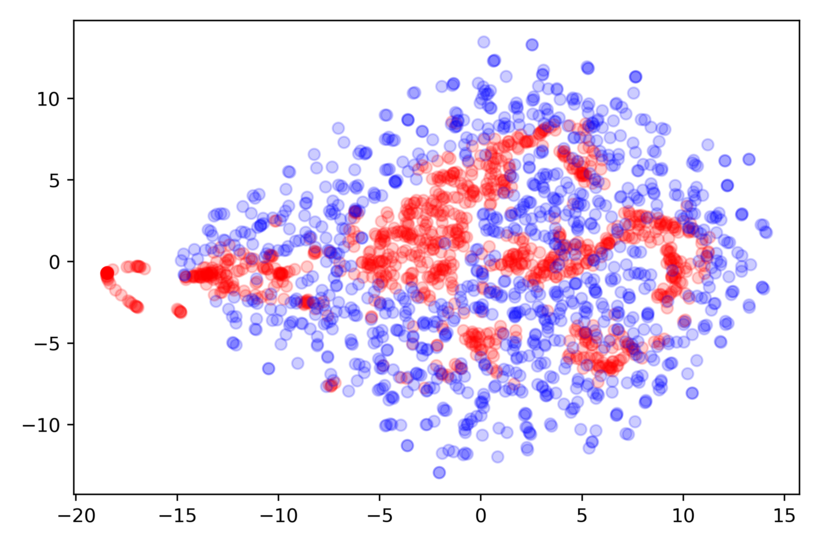}
        \caption{C-RNN-GAN}
    \end{subfigure}
    ~ 
    \begin{subfigure}[t]{0.14\textwidth}
        \centering
        \includegraphics[width=\textwidth]{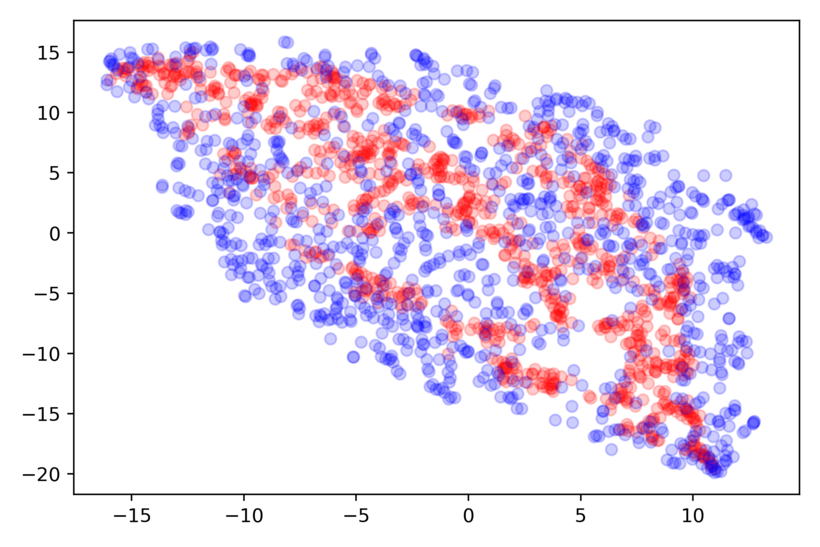}
        \includegraphics[width=\textwidth]{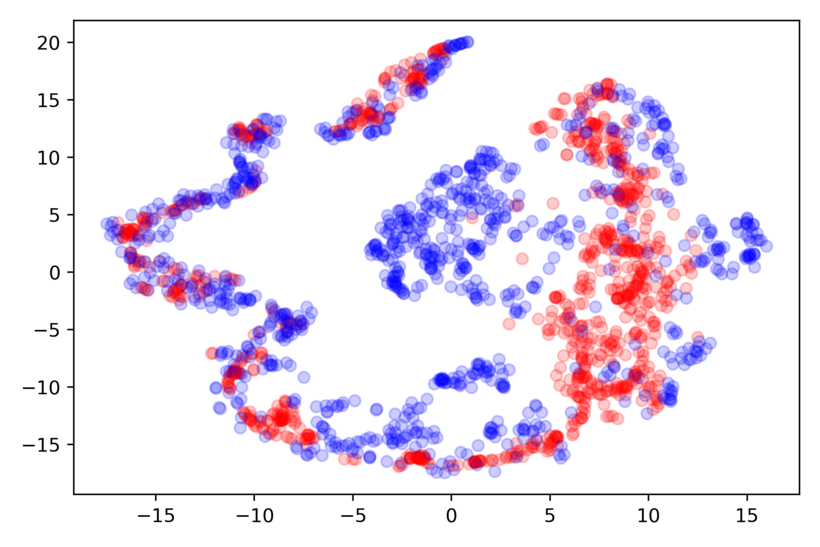}
        \includegraphics[width=\textwidth]{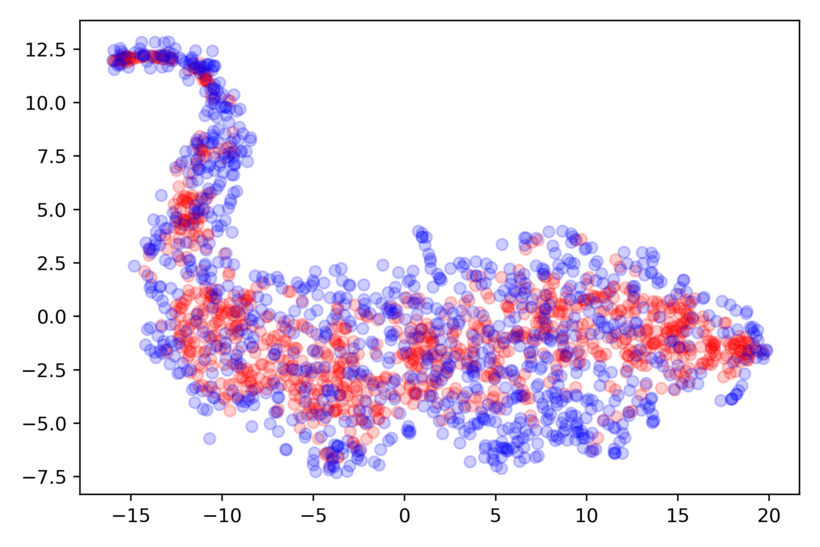}
        \includegraphics[width=\textwidth]{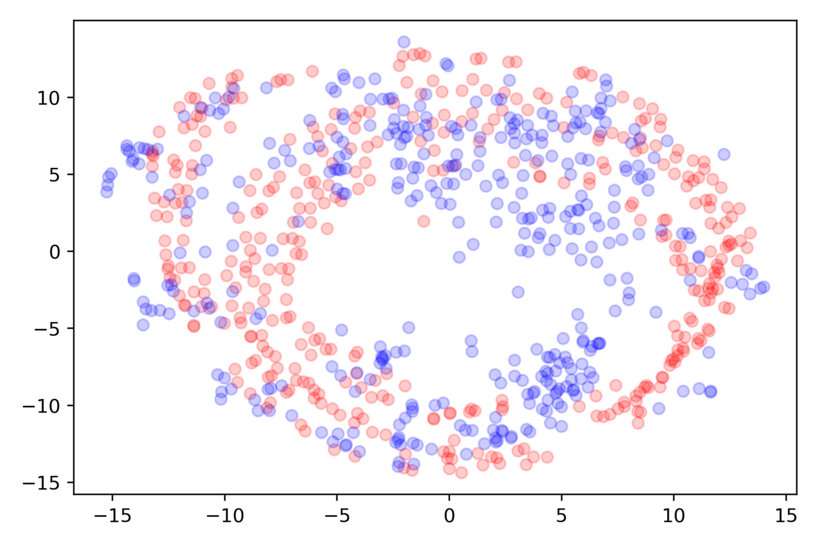}
        \includegraphics[width=\textwidth]{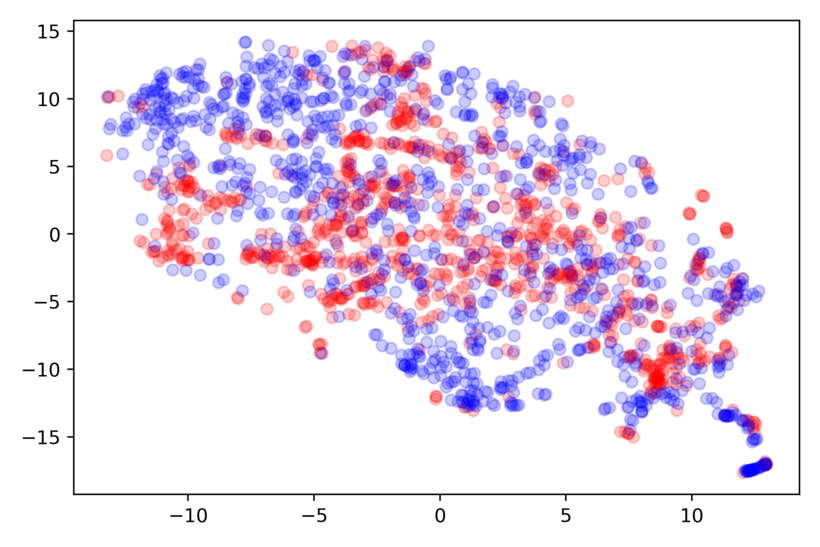}
        \caption{COT-GAN}
    \end{subfigure}
    ~ 
    \begin{subfigure}[t]{0.14\textwidth}
        \centering
        \includegraphics[width=\textwidth]{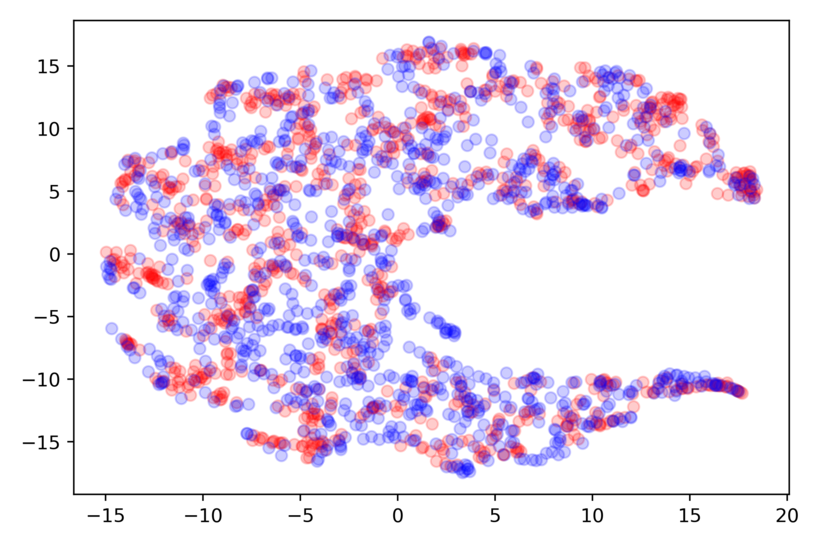}
        \includegraphics[width=\textwidth]{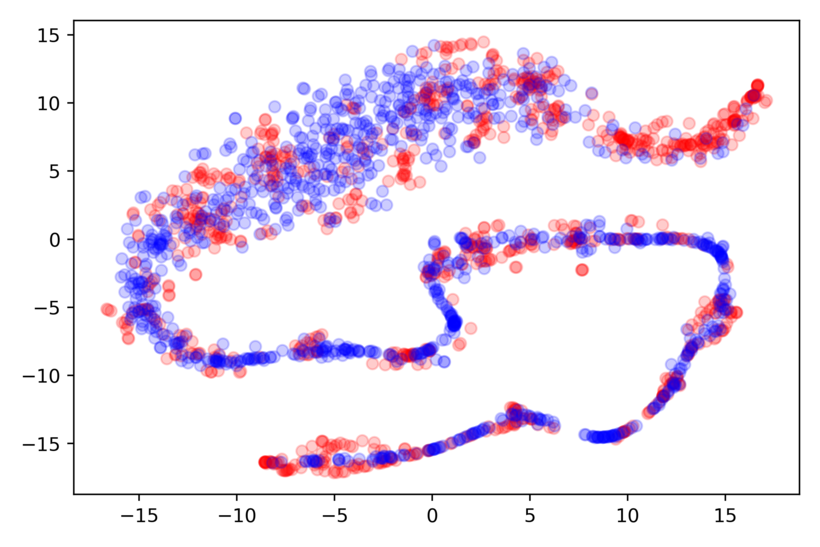}
        \includegraphics[width=\textwidth]{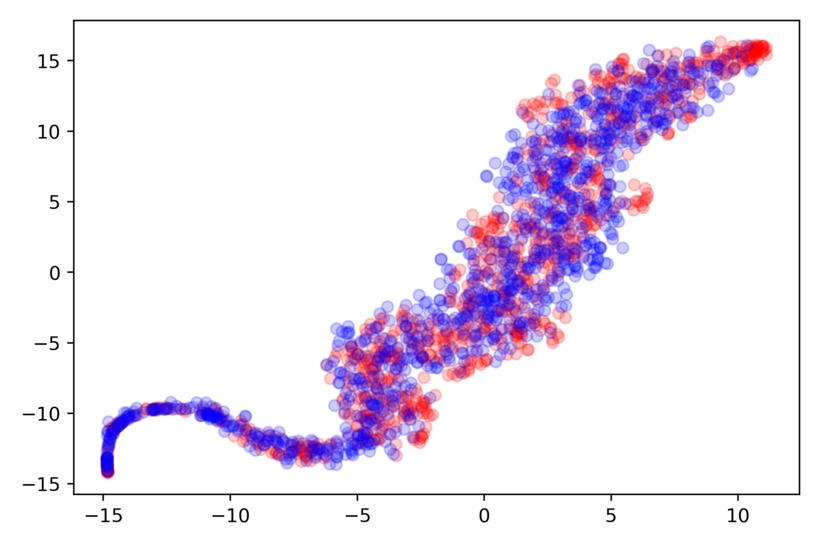}
        \includegraphics[width=\textwidth]{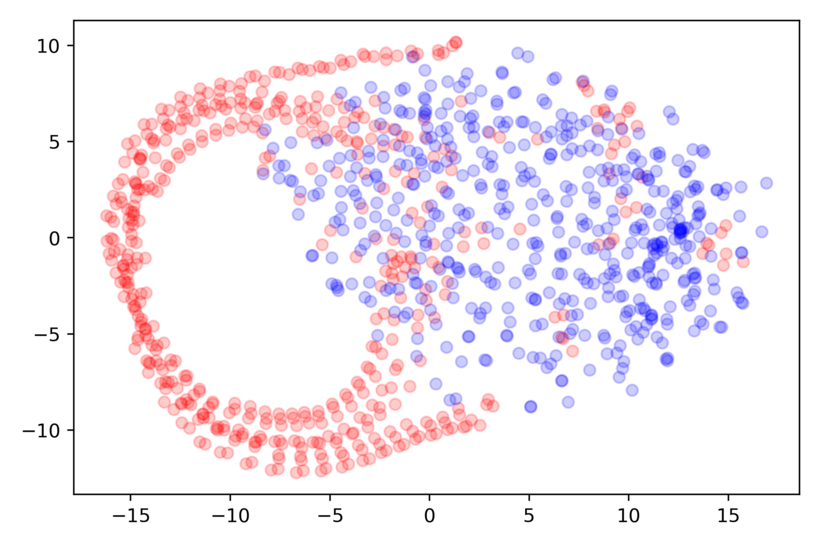}
        \includegraphics[width=\textwidth]{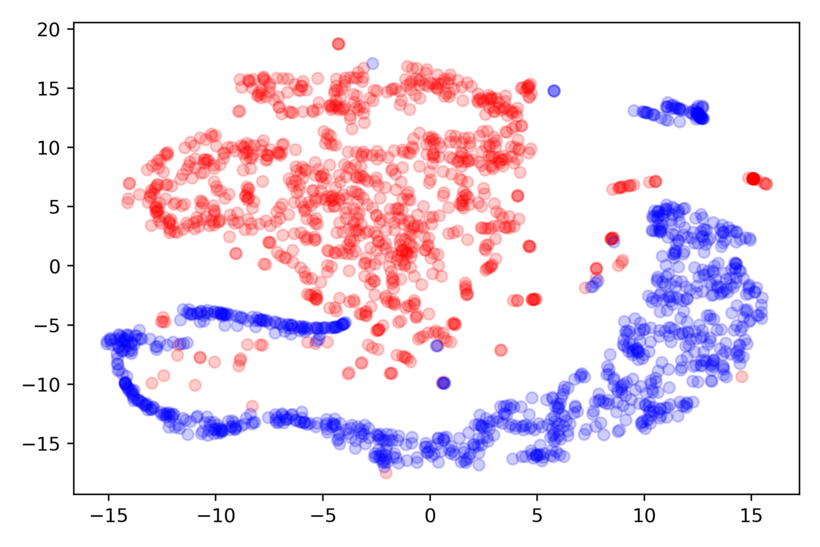}
        \caption{P-Forcing}
    \end{subfigure}

    \caption{t-SNE plots of Sines on the first row, Stocks on the second row, Energy on the third row, Chickenpox on the fourth row and Air on the fifth. Red indicates real data and blue indicates synthetic data. Best viewed in colour.}
    \label{fig:tsne plots}
\end{figure*}

We visualise t-SNEs in Figure~\ref{fig:tsne plots} and see that samples from TsT-GAN overlap real data samples extremely well for all datasets. The Chickenpox dataset is especially difficult to model due to the small number of samples, nevertheless TsT-GAN is able to achieve significant coverage. TimeGAN seems to learn specific modes, while also producing some out of distribution samples. RCGAN, C-RNN-GAN, COT-GAN and P-Forcing produce several out of distribution samples. COT-GAN particular produces impressive looking graphs, being commensurate with predictive scores. 

\begin{figure*}[t!]
    \centering
    
    \begin{subfigure}[t]{0.14\textwidth}
        \centering
        \includegraphics[width=\textwidth]{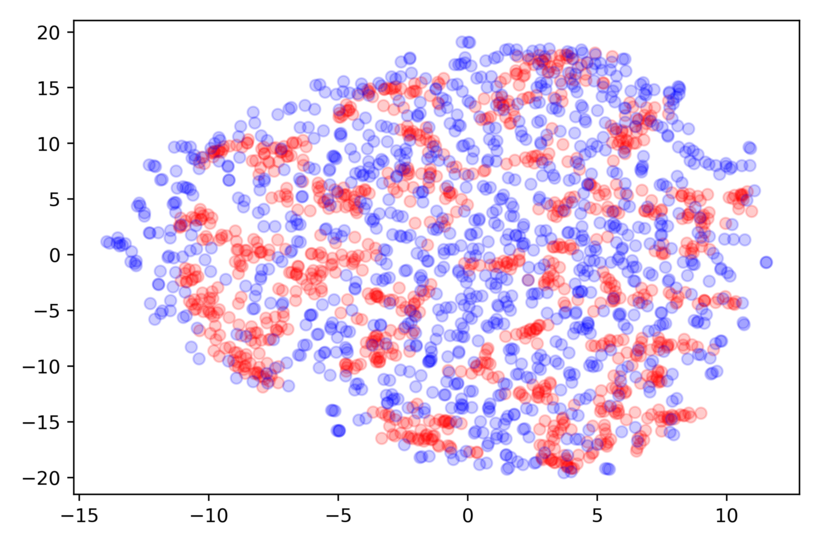}
        \includegraphics[width=\textwidth]{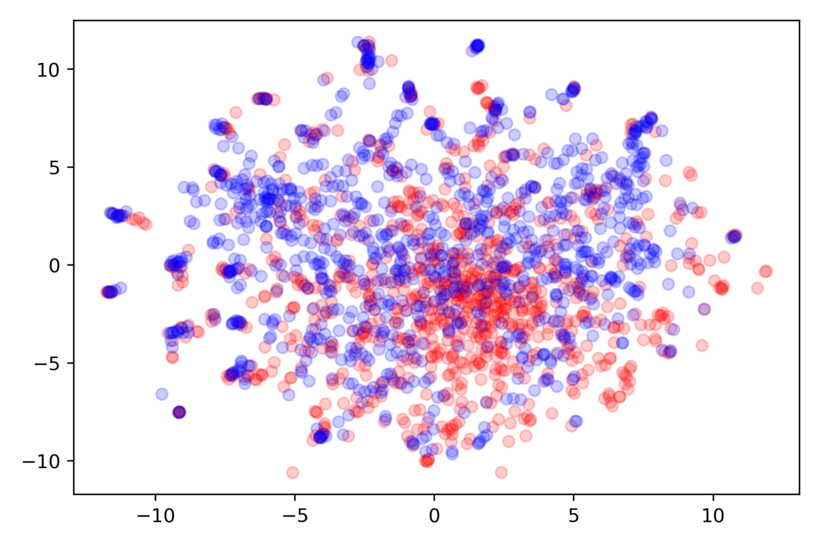}
        \includegraphics[width=\textwidth]{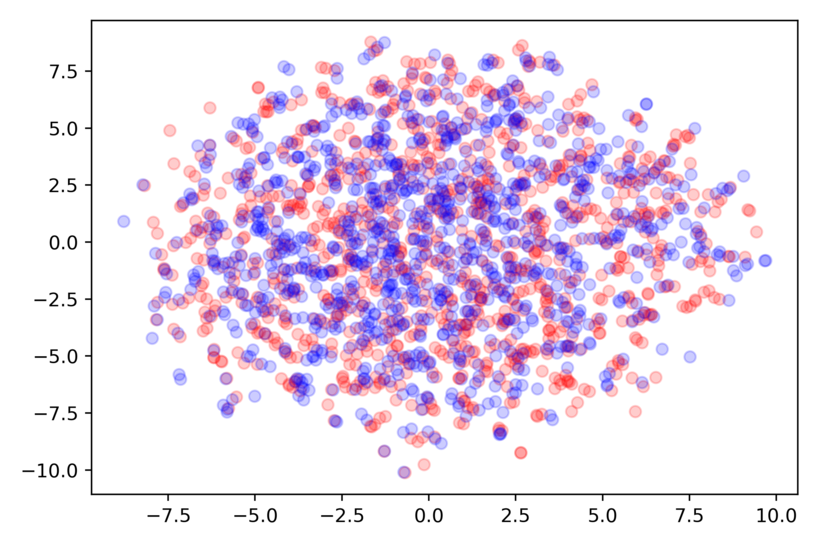}
        \includegraphics[width=\textwidth]{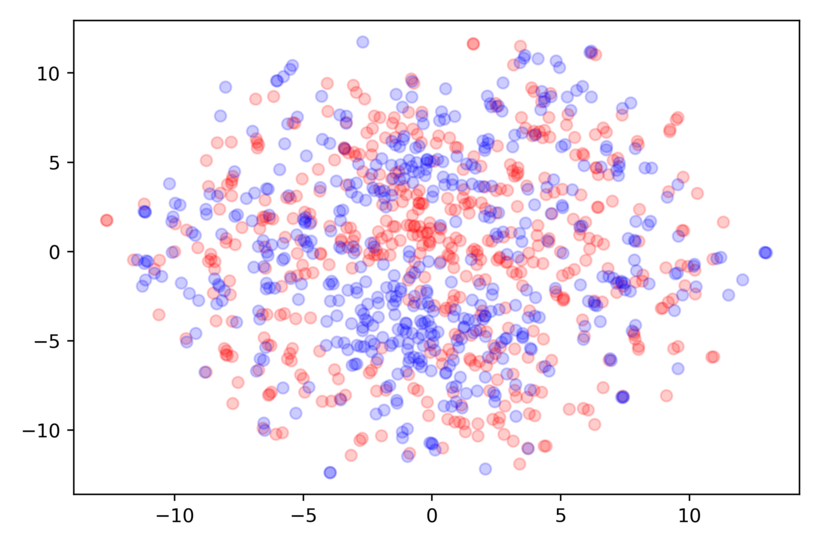}
        \includegraphics[width=\textwidth]{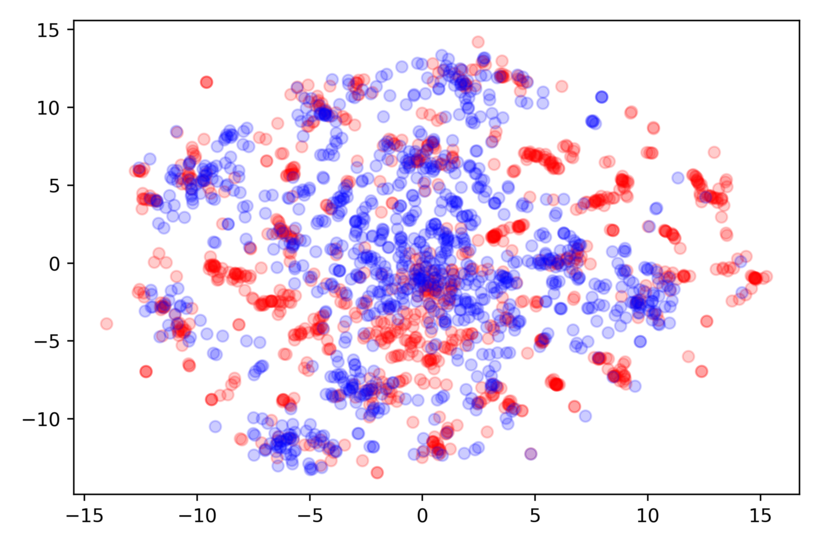}
        \caption{TsT-GAN}
    \end{subfigure}
    ~ 
    \begin{subfigure}[t]{0.14\textwidth}
        \centering
        \includegraphics[width=\textwidth]{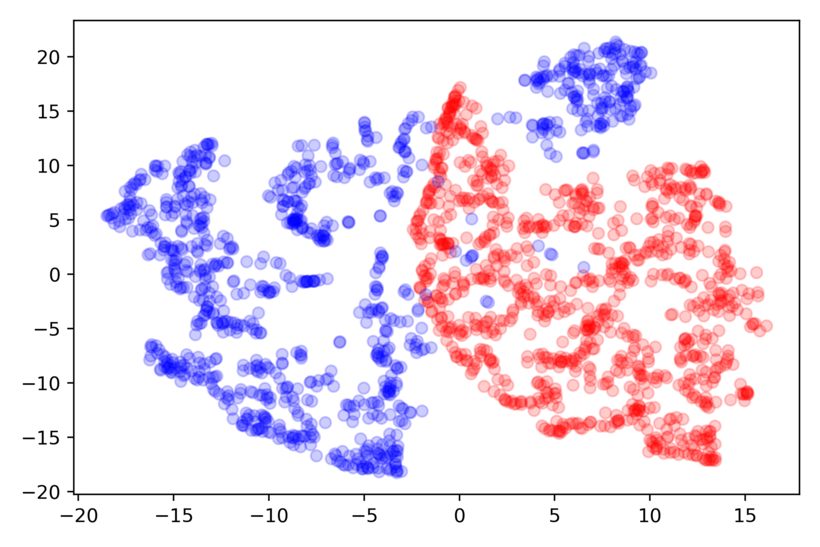}
        \includegraphics[width=\textwidth]{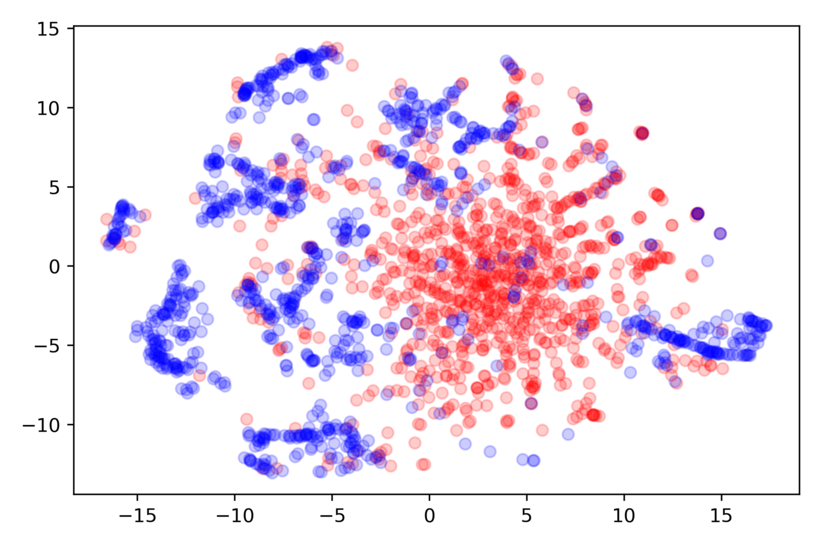}
        \includegraphics[width=\textwidth]{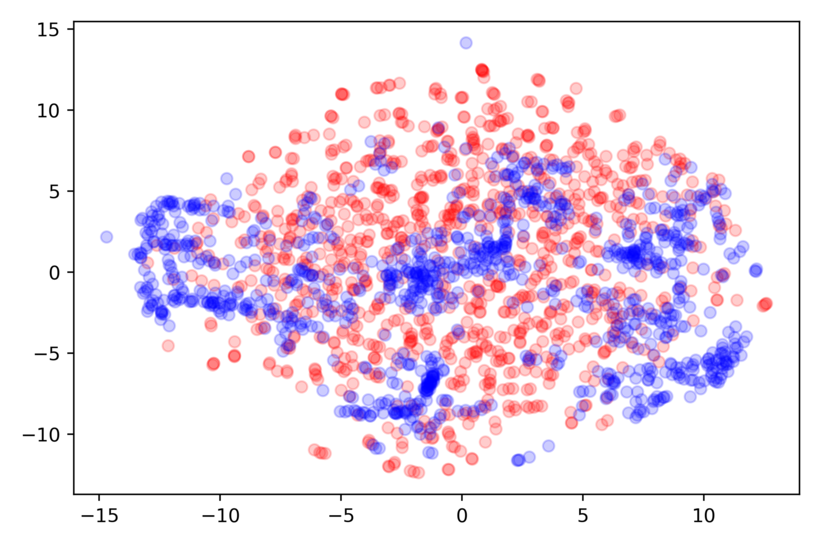}
        \includegraphics[width=\textwidth]{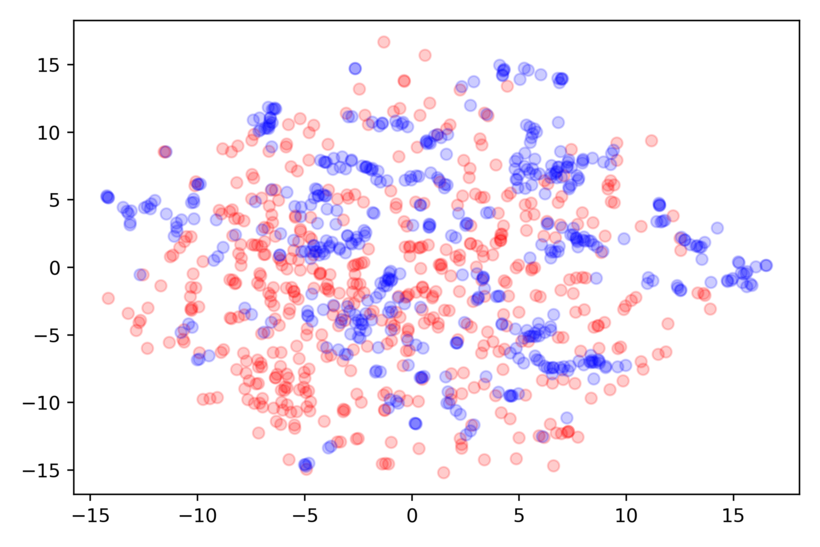}
        \includegraphics[width=\textwidth]{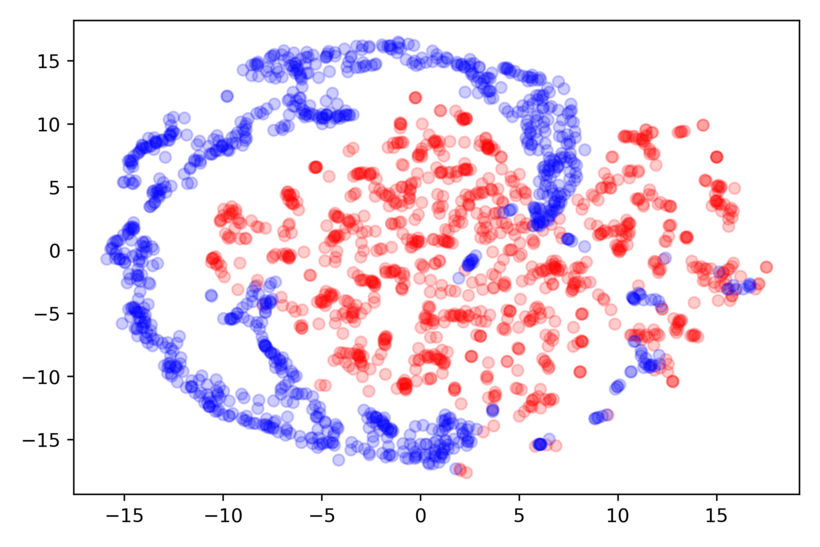}
        \caption{TimeGAN}
    \end{subfigure}
    ~
    \begin{subfigure}[t]{0.14\textwidth}
        \centering
        \includegraphics[width=\textwidth]{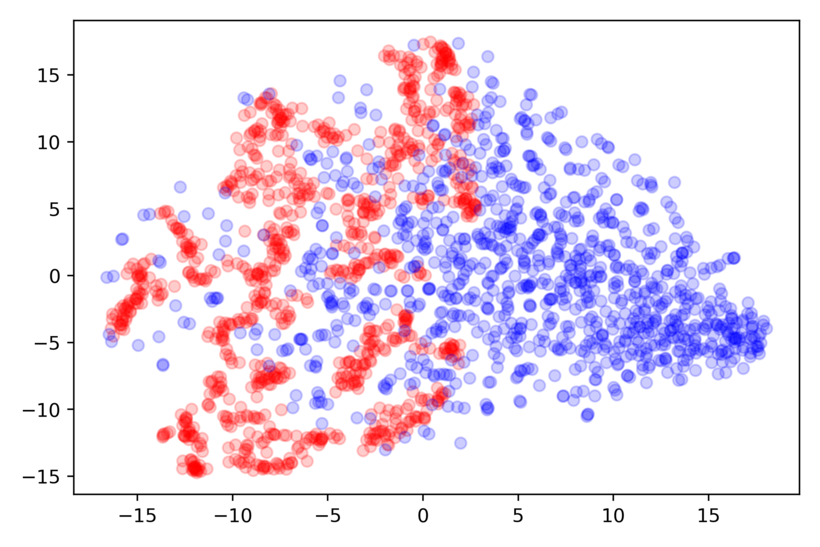}
        \includegraphics[width=\textwidth]{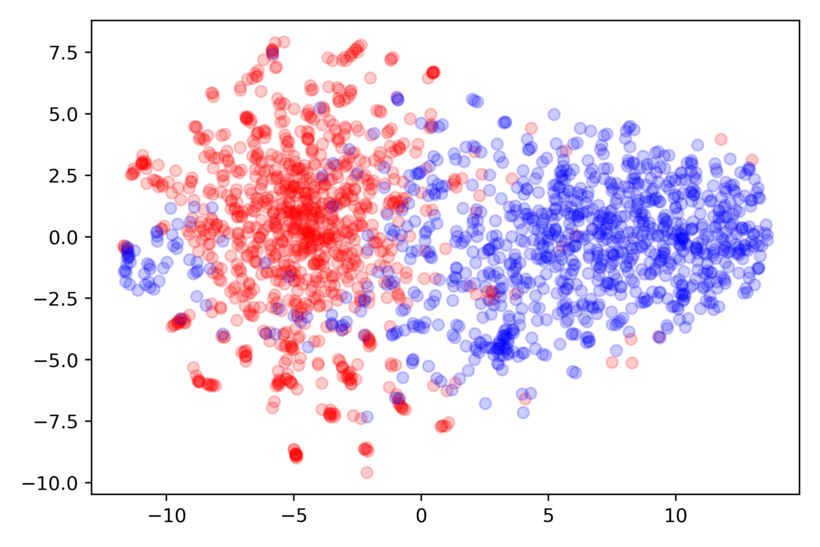}
        \includegraphics[width=\textwidth]{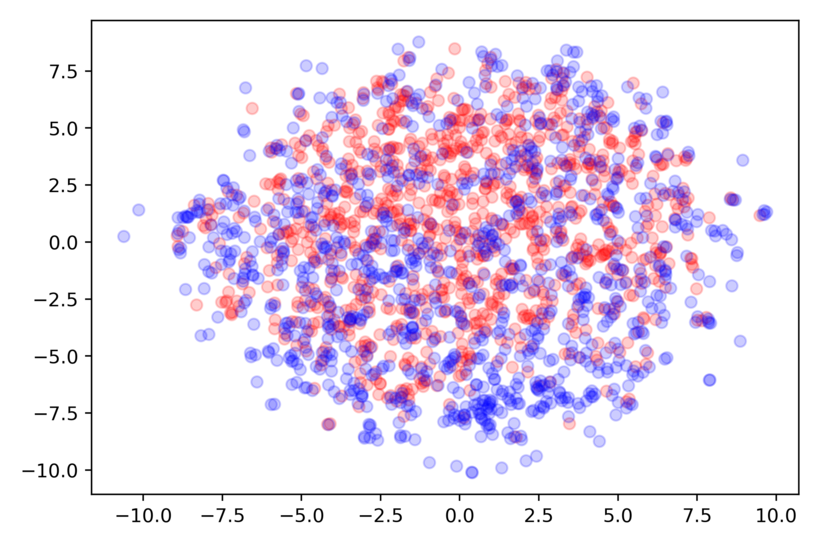}
        \includegraphics[width=\textwidth]{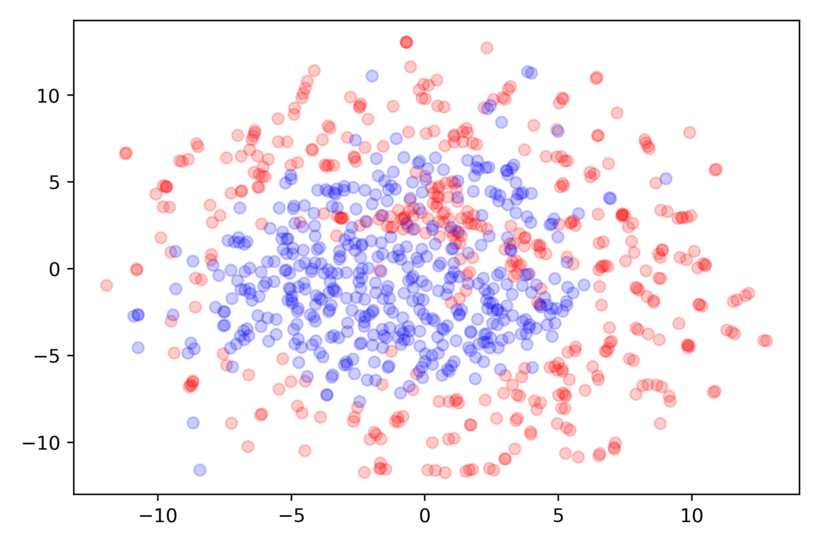}
        \includegraphics[width=\textwidth]{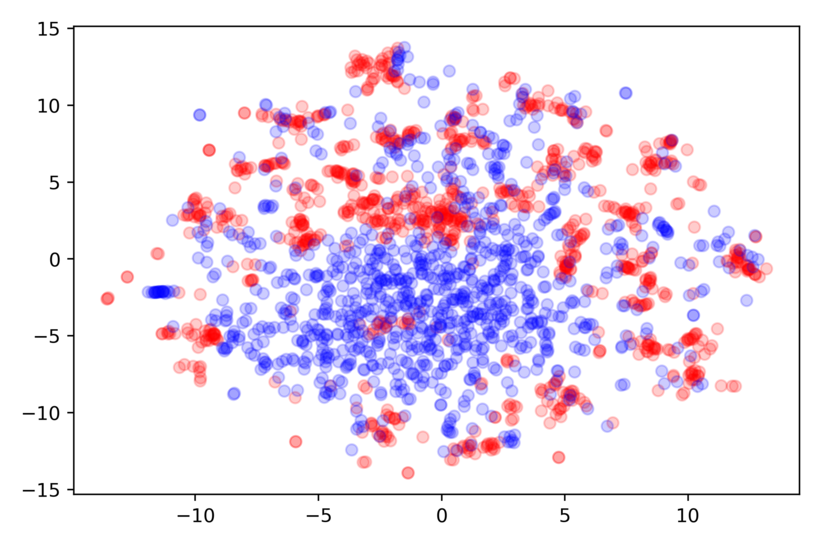}
        \caption{RCGAN}
    \end{subfigure}
    ~ 
    \begin{subfigure}[t]{0.14\textwidth}
        \centering
        \includegraphics[width=\textwidth]{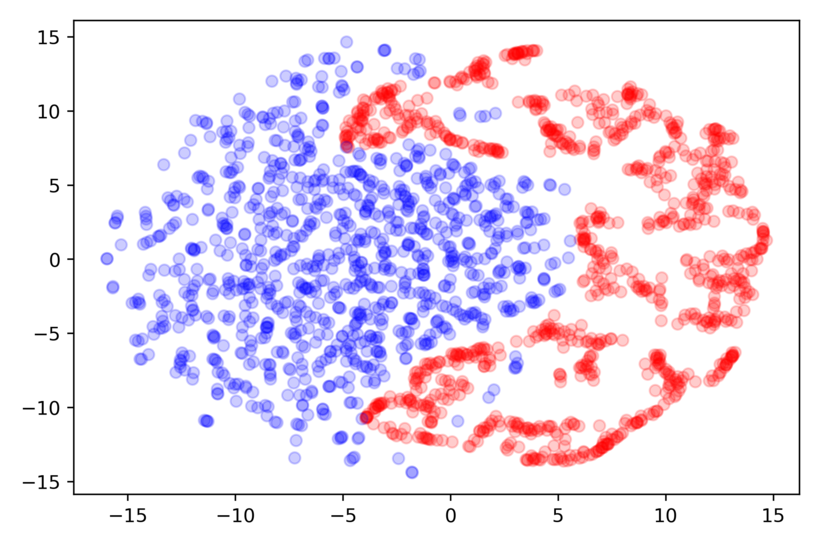}
        \includegraphics[width=\textwidth]{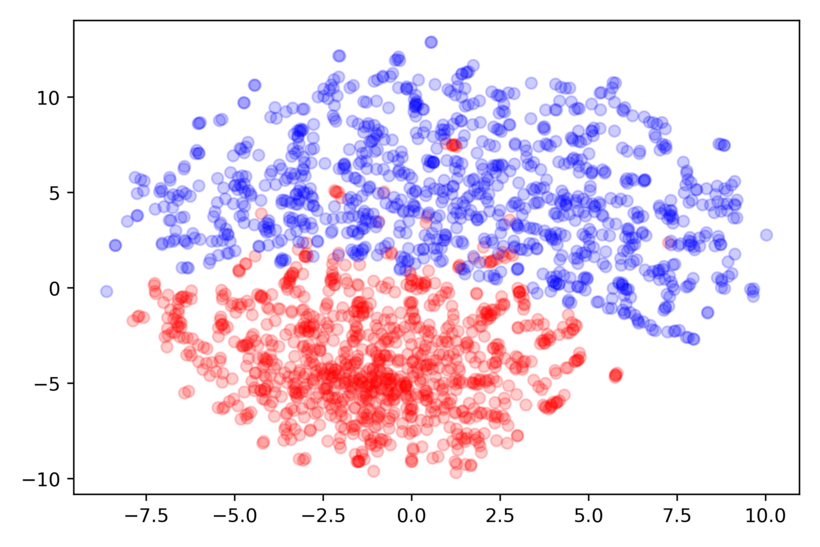}
        \includegraphics[width=\textwidth]{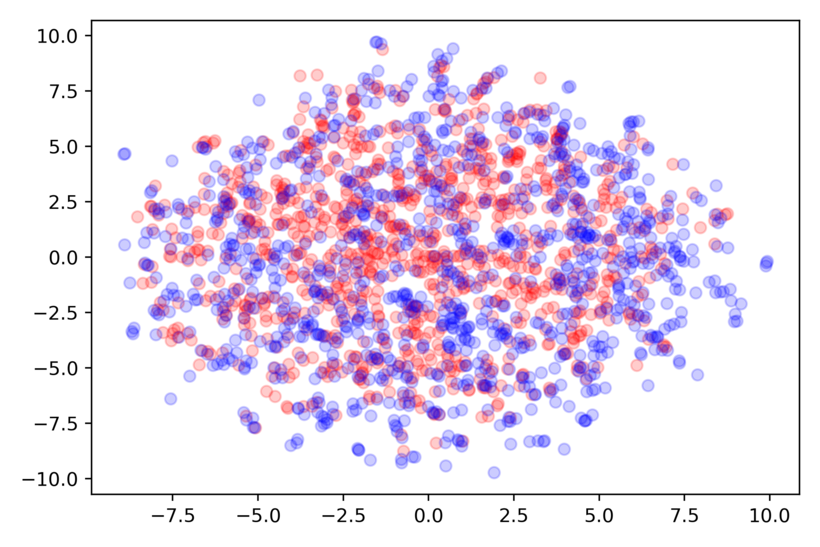}
        \includegraphics[width=\textwidth]{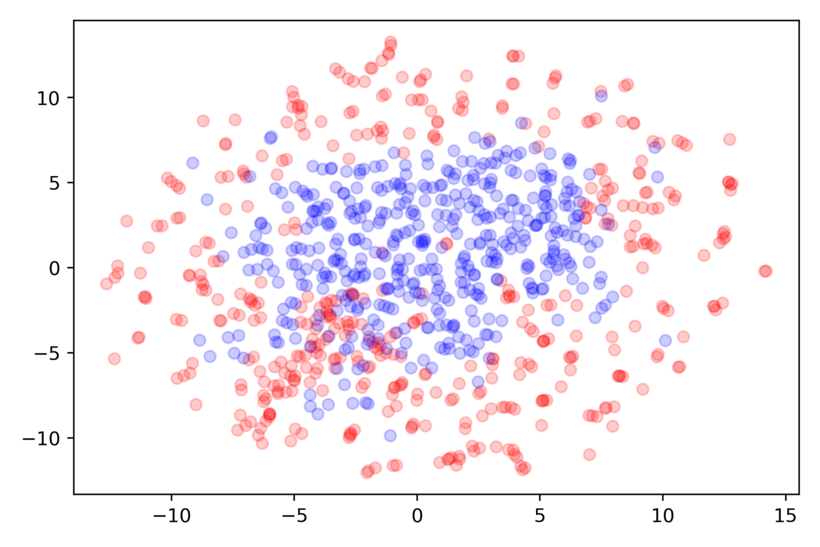}
        \includegraphics[width=\textwidth]{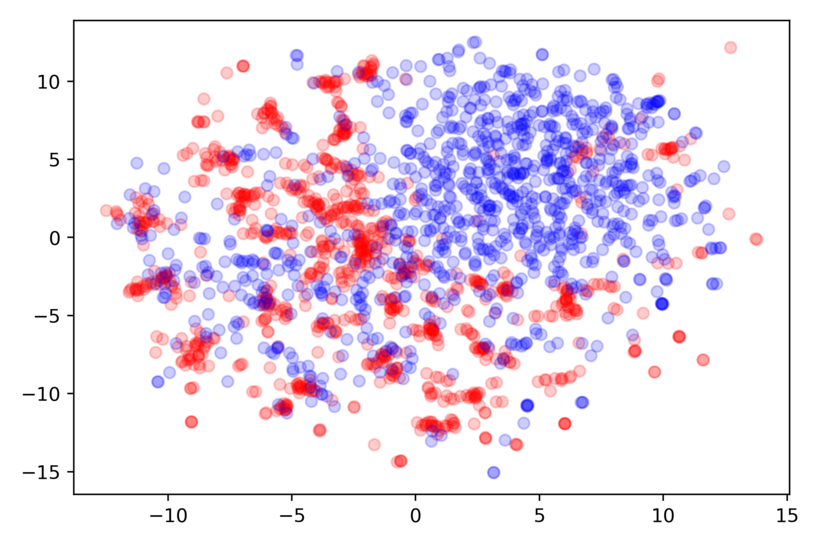}
        \caption{C-RNN-GAN}
    \end{subfigure}
    ~ 
    \begin{subfigure}[t]{0.14\textwidth}
        \centering
        \includegraphics[width=\textwidth]{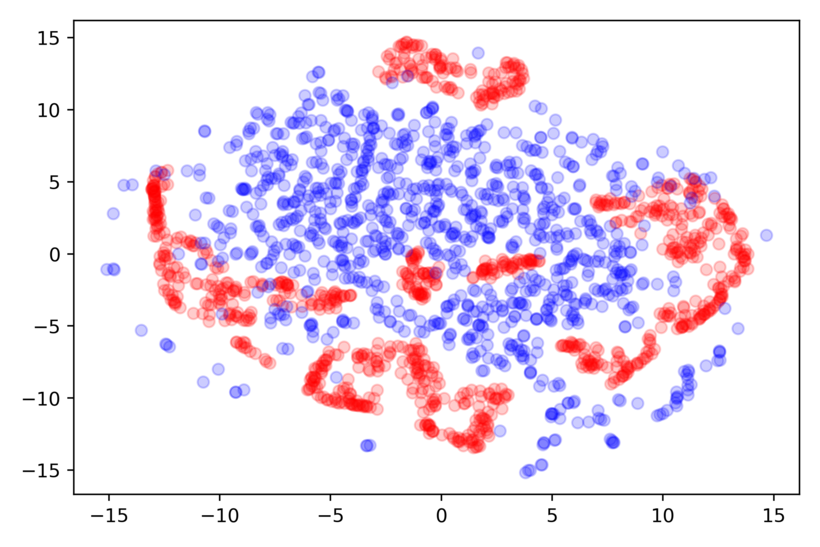}
        \includegraphics[width=\textwidth]{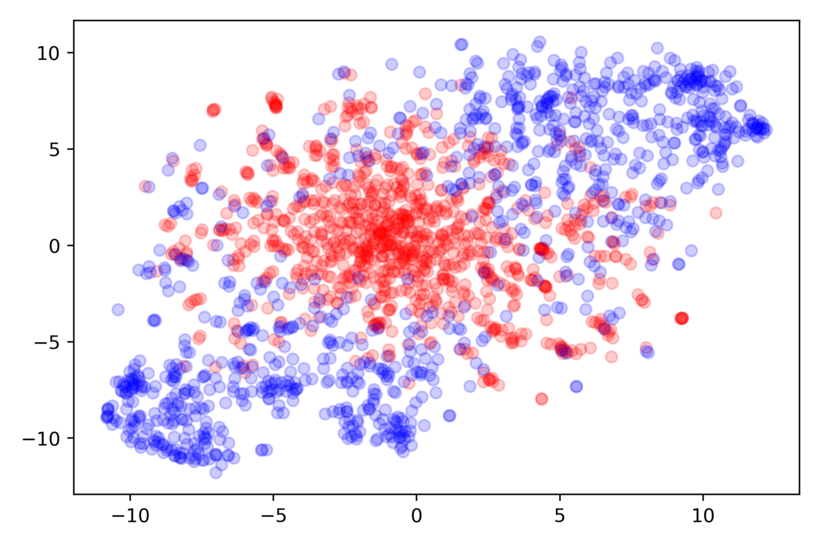}
        \includegraphics[width=\textwidth]{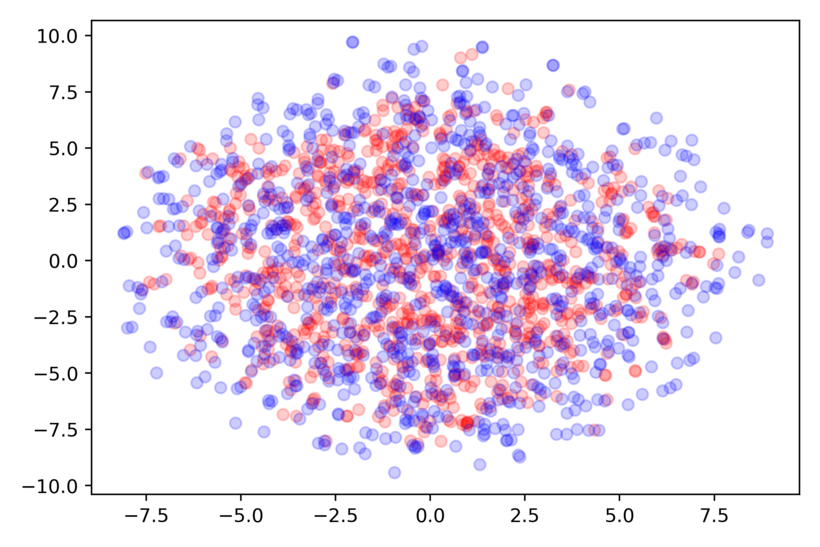}
        \includegraphics[width=\textwidth]{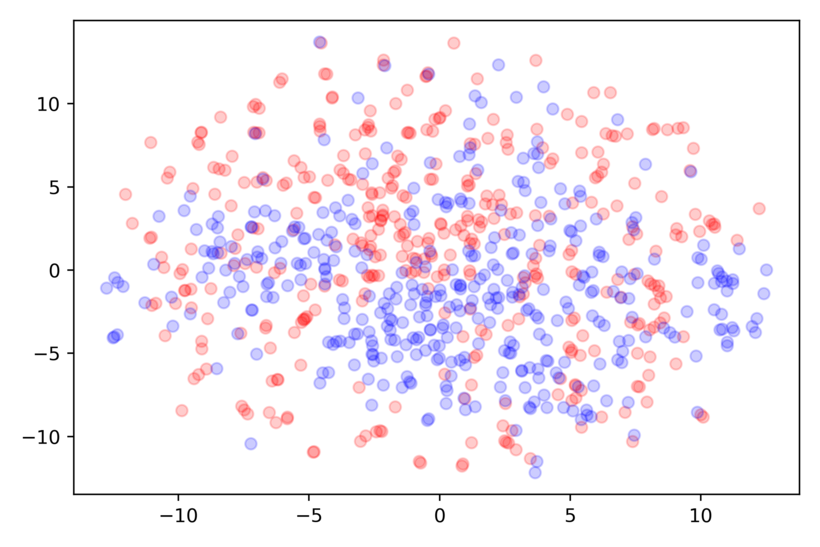}
        \includegraphics[width=\textwidth]{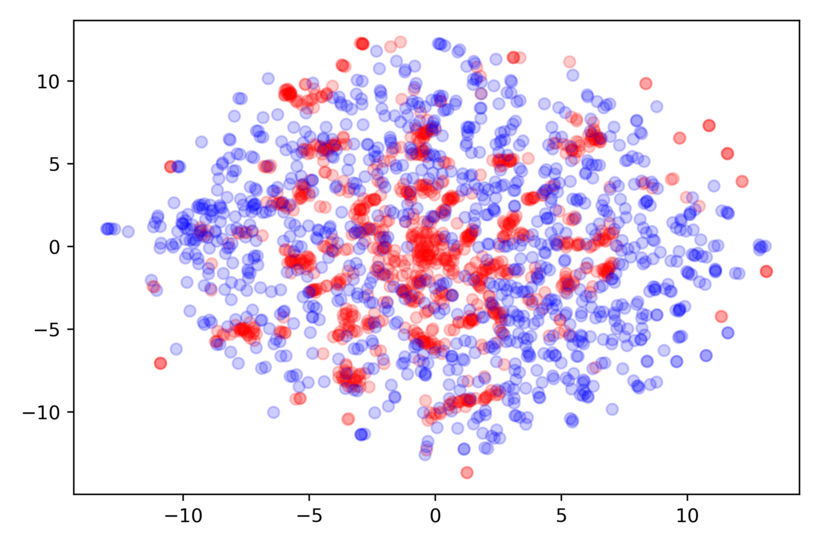}
        \caption{COT-GAN}
    \end{subfigure}
    ~ 
    \begin{subfigure}[t]{0.14\textwidth}
        \centering
        \includegraphics[width=\textwidth]{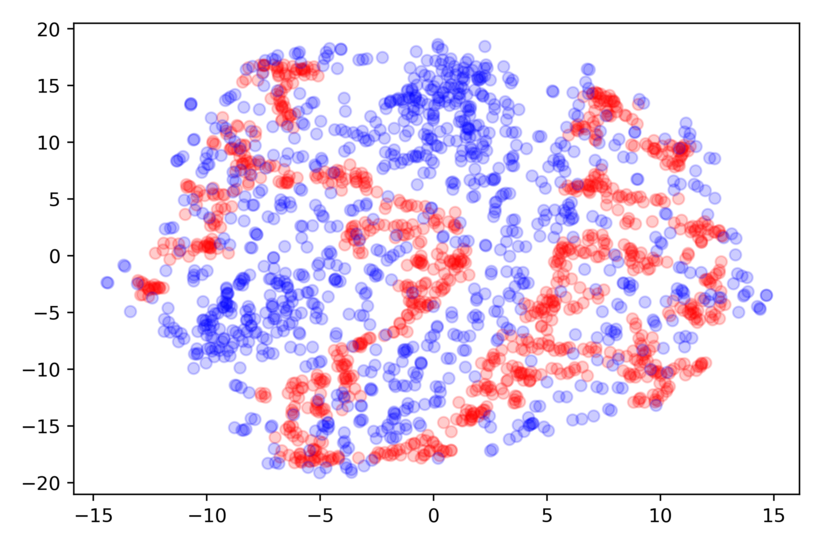}
        \includegraphics[width=\textwidth]{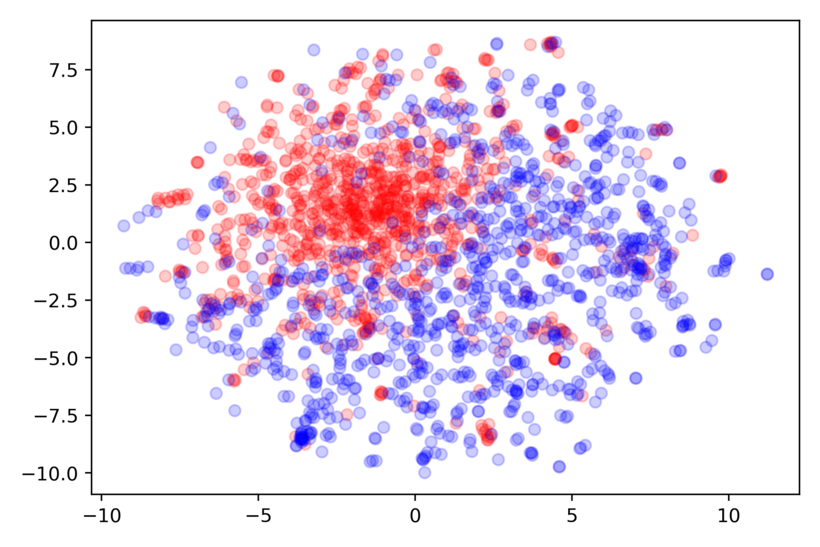}
        \includegraphics[width=\textwidth]{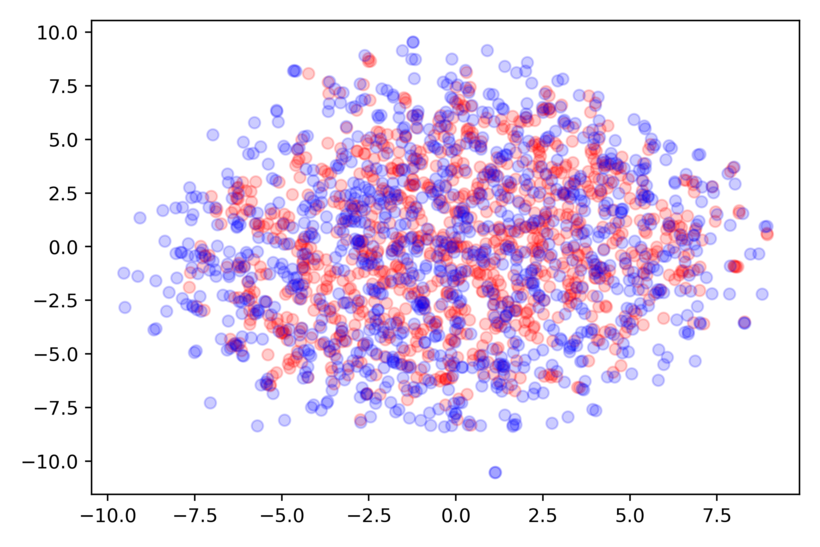}
        \includegraphics[width=\textwidth]{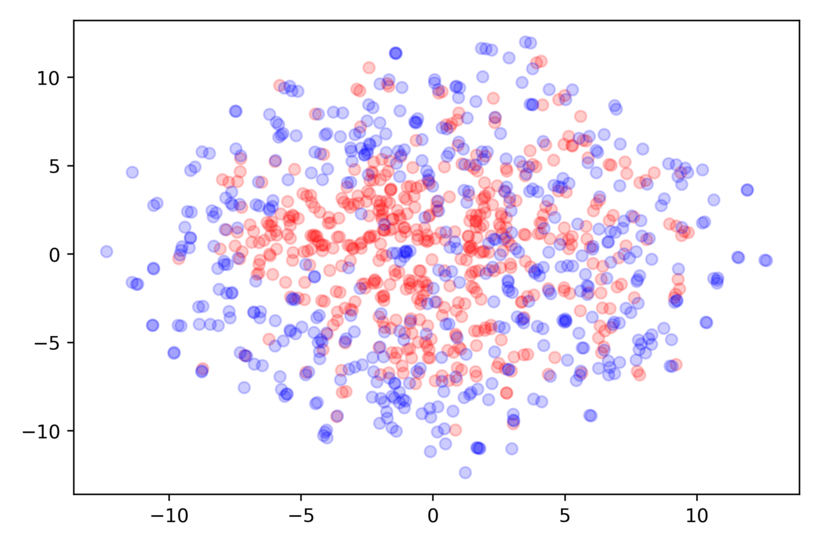}
        \includegraphics[width=\textwidth]{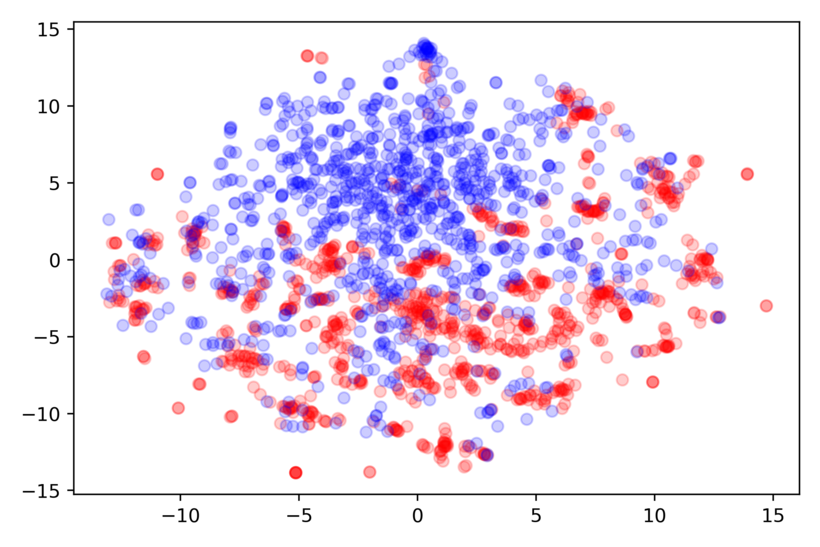}
        \caption{P-Forcing}
    \end{subfigure}

    \caption{t-SNE plots of first differences with Sines on the first row, Stocks on the second row, Energy on the third row, Chickenpox on the fourth row and Air on the fifth. Red indicates real data and blue indicates synthetic data. Best viewed in colour.}
    \label{fig:tsne first diff plots}
\end{figure*}

From Figure~\ref{fig:tsne first diff plots} we see that TsT-GAN captured the first differences well in all datasets. Of particular interest is the Sines row. Sines is a toy dataset with known generating distributions where first differences follow specific patterns (the differences are themselves sinusoidal); had TsT-GAN learned fully the generating distribution, we would expect to see distinct regions of high and low density for synthetic samples overlapping the true samples. 

\begin{table}
  \caption{Predictive and Discriminative scores with standard deviations for both comparison with baselines and ablations. Lower scores are better and best performance is indicated in bold. Predictive scores include score on the original data for comparison.}
  \label{tab:predictive discriminative scores}
  \centering
  \begin{tabular}{c|l|l|l|l|l}
    \toprule
    \multicolumn{6}{c}{\textbf{Predictive Score (MAE) }} \\
    \toprule
    
    Model & Sines & Stocks & Energy & Chickenpox & Air \\ \hline
    Original  & .009 $\pm$  .000 & .010 $\pm$  .001 & .032 $\pm$  .001 & .089 $\pm$  .002 & .034 $\pm$ .001 \\ \hdashline
    
    TsT-GAN & \textbf{.008 $\pm$ .000} & \textbf{.009 $\pm$ .000} & \textbf{.039 $\pm$ .001} & \textbf{.091 $\pm$ .001} & \textbf{.042 $\pm$ .002} \\ \hline
    
    TimeGAN & .024 $\pm$ .004 & .011 $\pm$  .001 & .050 $\pm$ .001 & .101 $\pm$  .002 & .114 $\pm$ .005 \\ \hline
    
    RCGAN & .012 $\pm$ .000 & .021 $\pm$ .001 & .068 $\pm$ .001 & .106 $\pm$ .002 & .072 $\pm$ .001 \\ \hline
    
    C-RNN-GAN & .017 $\pm$ .000 & .027 $\pm$ .001 & .069 $\pm$ .001 & .207 $\pm$ .002 & .095 $\pm$ .002 \\ \hline
    
    COT-GAN & .016 $\pm$ .000 & .012 $\pm$  .000 & .056 $\pm$ .001 & .094 $\pm$ .003 & .044 $\pm$ .000 \\ \hline 
    
    P-Forcing & .014 $\pm$ .001 & .018 $\pm$ .000 & .059 $\pm$  .003 & .319 $\pm$ .002 & .190 $\pm$ .041 \\
    
    \end{tabular}
    \begin{tabular}{c|l|l|l|l|l}
    \toprule
    \multicolumn{6}{c}{\textbf{Discriminative Score (proportion classified as synthetic)}} \\
    \toprule
    
    TsT-GAN & \textbf{.026 $\pm$ .018} & .122 $\pm$ .020 & \textbf{.442 $\pm$  .007} & .053 $\pm$  .044 & \textbf{.243 $\pm$ .009} \\ \hline
    
    TimeGAN & .231 $\pm$ .117 & .191 $\pm$ .029 & .496 $\pm$ .004 & .046 $\pm$ .044 & .479 $\pm$ .025 \\ \hline
    
    RCGAN & .174 $\pm$ .033 & .260 $\pm$ .010 & .500 $\pm$ .000 & .050 $\pm$ .003 & .478 $\pm$ .004 \\ \hline
    
    C-RNN-GAN & .274 $\pm$ .040 & .290 $\pm$ .032 & .547 $\pm$ .004 & .209 $\pm$ .001 & .473 $\pm$ .024 \\ \hline
    
    COT-GAN & .302 $\pm$ .089 & .260 $\pm$ .068 & .500 $\pm$ .000 & \textbf{.006 $\pm$ .053} & .441 $\pm$ .052\\ \hline
    
    P-Forcing & .253 $\pm$ .036 & \textbf{.088 $\pm$ .007} & .553 $\pm$ .044 & .587 $\pm$ .009 & .484 $\pm$ .018 \\
    
    \end{tabular}
    \begin{tabular}{c|l|l|l|l|l}
    \toprule
    \multicolumn{6}{c}{\textbf{Ablations Predictive Score (MAE)}} \\
    \toprule
    
    TsT-GAN & \textbf{.008 $\pm$ .000} & \textbf{.009 $\pm$ .000} & \textbf{.039 $\pm$ .001} & \textbf{.091 $\pm$ .001} & \textbf{.042 $\pm$ .002} \\ \hline
    
    - ML & \textbf{.008 $\pm$ .007} & .011 $\pm$  .001 & .047 $\pm$ .006 & .101 $\pm$  .002 & .045 $\pm$ .002 \\ \hline
    
    - MM + Auto & \textbf{.008 $\pm$ .001} & .012 $\pm$  .000 & .054 $\pm$ .001 & .111 $\pm$  .002 & .046 $\pm$ .001 \\ \hline
    
    - Embedding & .009 $\pm$ .000 & .016 $\pm$ .000 & .081 $\pm$ .001 & .145 $\pm$ .004 & .056 $\pm$ .003\\ \hline
    
    - MM & .009 $\pm$ .001 & .013 $\pm$ .001 & .057 $\pm$ .001 & .095 $\pm$ .002 & .051 $\pm$ .002\\ \hline
    
    Base & .010 $\pm$  .001 & .020 $\pm$ .000 & .089 $\pm$ .001 & .196 $\pm$ .006 & .068 $\pm$ .003 \\ 
    
    \end{tabular}
    \begin{tabular}{c|l|l|l|l|l}
    \toprule
    \multicolumn{6}{c}{\textbf{Ablations Discriminative Score (proportion classified as synthetic)}} \\
    \toprule
    
    TsT-GAN & \textbf{.026 $\pm$ .018} & .122 $\pm$ .020 & \textbf{.442 $\pm$  .007} & .053 $\pm$  .044 & \textbf{.243 $\pm$ .009} \\ \hline
    
    - ML & .028 $\pm$ .010 & .140 $\pm$  .092 & .465 $\pm$ .003 & .069 $\pm$ .031 & .302 $\pm$ .003\\ \hline
    
    - MM + Auto & .029 $\pm$ .011 & \textbf{.118 $\pm$  .089} & .488 $\pm$ .004 & \textbf{.048 $\pm$ .017} & .452 $\pm$ .004 \\ \hline

    - Embedding & .145 $\pm$ .079 & .254 $\pm$ .032 & .497 $\pm$ .003 & .342 $\pm$ .030 & .514 $\pm$ .005 \\ \hline
    
    - MM & .166 $\pm$ .047 & .171 $\pm$ .095 & .498 $\pm$ .001 & .480 $\pm$ .069 & .477 $\pm$ .002 \\ \hline
    
    Base & .113 $\pm$ .042 & .200 $\pm$ .035 & .529 $\pm$ .0120 & .462 $\pm$ .126 & .613 $\pm$ .015 \\
    
    \bottomrule
  \end{tabular}
\end{table}

\subsection{Ablation Experiments}

We perform ablation experiments to evaluate the importance of each component of TsT-GAN. Our experiments are as follows:

\begin{itemize}
    \item \textbf{- ML} Removes only the auxiliary moment matching objective. All subsequent ablations remove the moment matching objective as well. 
    
    \item \textbf{- MM + Auto} Makes the generator autoregressive and removes the masked modelling objective, replacing it with a one step ahead prediction objective.
    
    \item \textbf{- Embedding} Removes the embedding network resulting in a generator that is trained with the LS-GAN objective and MM loss, with the parameters of the predictor network being updated jointly with the generator. 
    
    \item \textbf{- MM} Removes only the masked modelling objective but retains the bidirectional generator.
    
    \item \textbf{Base} Is a standard transformer GAN made by removing MM and the embedding network. 
\end{itemize}

From the latter two sections in Table~\ref{tab:predictive discriminative scores} we see the TsT-GAN outperforming all ablations. The autoregressive generator outperforms TsT-GAN in the discriminative score for Stocks and Chickenpox, although the difference is small. Removing the embedding network in particular has a a significant detrimental effect on predictive performance performance on all but the Sines dataset, suggesting that our enforcement of the conditional distribution plays an important role in capturing useful temporal correlations across time.

\section{Conclusion}

We have presented TsT-GAN, a new framework for training time-series generative models. The unconditional generator network in TsT-GAN is guided by unsupervised masked modelling to produce high quality synthetic sequences that capture both the global distribution as well as conditional time-series dynamics. We evaluate and benchmark our model using the TS-TR framework and show that TsT-GAN consistently outperforms existing methods. Future work could explore how to better incorporate moment matching in a unified framework, rather than as an auxiliary loss. Furthermore, TsT-GAN's discriminative scores still show scope for improvement suggesting that there still exists some discrepancy between true and learned distributions. A major limitation of our model is architecturally rooted: the Transformer architecture's self-attention mechanism has a computational complexity of $O(N^2)$ for a sequence of length $N$. As three out four components of TsT-GAN consist of Transformers, this results in a significant computational cost when training and performing inference with longer time-series. 

Our model can also contribute to data compression. As data increases in resolution and demand for data increases, it is crucial to ensure that data remains accessible. We have shown that our model is able to learn meaningful representations of several time-series datasets as well as its utility in downstream tasks. In future applications, a trained model could be disseminated instead of a much larger dataset.

%\newpage
\bibliographystyle{plain} % We choose the "plain" reference style
\bibliography{mybib}

%%%%%%%%%%%%%%%%%%%%%%%%%%%%%%%%%%%%%%%%%%%%%%%%%%%%%%%%%%%%

\end{document}